\documentclass{article}

\usepackage[preprint]{neurips_2024}
\usepackage{amsmath}
\usepackage{adjustbox}

\usepackage{algorithm}
\usepackage{algpseudocode}

\usepackage[utf8]{inputenc} %
\usepackage[T1]{fontenc}    %
\usepackage[colorlinks=true, linkcolor=blue, urlcolor=blue, citecolor=blue]{hyperref}       %
\usepackage{url}            %
\usepackage{booktabs}       %
\usepackage{amsfonts}       %
\usepackage{nicefrac}       %
\usepackage{microtype}      %
\usepackage{xcolor}         %
\usepackage{graphicx}

\usepackage{ragged2e}

\title{Stochastic Parameter Decomposition}

\author{%
  Lucius Bushnaq\thanks{Goodfire -- Work primarily carried out while at Apollo Research} \\
  \texttt{lucius@goodfire.ai} \\
\And
  Dan Braun\footnotemark[1]  \\
  \texttt{dan.braun@goodfire.ai} \\
\And
  Lee Sharkey\thanks{Goodfire} \\
  \texttt{lee@goodfire.ai} \\
}

\begin{document}

\maketitle
\setcounter{footnote}{0}

\begin{abstract}

A key step in reverse engineering neural networks is to decompose them into simpler parts that can be studied in relative isolation. Linear parameter decomposition---a framework that has been proposed to resolve several issues with current decomposition methods---decomposes neural network parameters into a sum of sparsely used vectors in parameter space. However, the current main method in this framework, Attribution-based Parameter Decomposition (APD), is impractical on account of its computational cost and sensitivity to hyperparameters. In this work, we introduce \textit{Stochastic Parameter Decomposition} (SPD), a method that is more scalable and robust to hyperparameters than APD, which we demonstrate by decomposing models that are slightly larger and more complex than was possible to decompose with APD. We also show that SPD avoids other issues, such as shrinkage of the learned parameters, and better identifies ground truth mechanisms in toy models. By bridging causal mediation analysis and network decomposition methods, this demonstration opens up new research possibilities in mechanistic interpretability by removing barriers to scaling linear parameter decomposition methods to larger models. We release a library for running SPD and reproducing our experiments at \url{https://github.com/goodfire-ai/spd/tree/spd-paper}.

\end{abstract}

\section{Introduction}

We have little understanding of the internal mechanisms that neural networks learn that enable their impressive capabilities. Understanding---or reverse engineering---these mechanisms may enable us to better predict and design neural network behavior and propensities for the purposes of safety and control. It may also be useful for scientific knowledge discovery: Neural networks can often perform better than humans on some tasks. They must therefore `know' things about the world that we do not know---things that we could uncover by understanding their mechanisms. 

An important first step to reverse engineering neural networks is to decompose them into individual mechanisms whose structure and interactions can be studied in relative isolation. Previous work has taken a variety of approaches to network decomposition. A popular approach is sparse dictionary learning (SDL) \citep{cunningham2023sparseautoencodershighlyinterpretable, bricken2023monosemanticity}, which aims to decompose neural network activations by optimizing sparsely activating dictionary elements to reconstruct or predict neural activation vectors. However, this approach suffers from a range of conceptual and practical problems, such as failing to account for feature geometry \citep{leask2025sparseautoencoderscanonicalunits, mendel2024sae} and not decomposing networks into functional components \citep{chanin2024absorptionstudyingfeaturesplitting, bricken2023monosemanticity, till2024truefeatures} (see \citet{sharkey2025openproblemsmechanisticinterpretability} for a review). 

Recently, \textit{linear parameter decomposition} \citep{braun2025interpretabilityparameterspaceminimizing}, has been proposed to address some of the issues faced by SDL and other current approaches. Instead of decomposing networks into directions in activation space, linear parameter decomposition methods decompose networks into \textit{vectors in parameter space}, called \textit{parameter components}. Parameter components are selected such that, simultaneously, (a) they sum to the parameters of the original model, (b) as few as possible are required to replicate the network's behavior on any given input, and (c) they are as `simple' as possible. This approach promises a framework that suggests solutions to issues like `feature splitting' \citep{chanin2024absorptionstudyingfeaturesplitting, bricken2023monosemanticity}; the foundational conceptual issue of defining a `feature' (by re-basing it in the language of `mechanisms'); and the issues of multidimensional features and feature geometry \citep{braun2025interpretabilityparameterspaceminimizing}. It also suggests a new way to bridge mechanistic interpretability and causal mediation analysis \citep{mueller2024questrightmediatorhistory}.

However, \textit{Attribution-based Parameter Decomposition} (APD) \citep{braun2025interpretabilityparameterspaceminimizing}, the only method that has been so far proposed for linear parameter decomposition (with which this paper assumes some familiarity), suffers from several significant issues that hinder its use in practice, including: 
\begin{enumerate}
    \item \textbf{Scalability}: APD has a high memory cost, since it decomposes a network into many parameter components, each of which is a whole vector in parameter space. They therefore each have the same memory cost as the original network. 
    \item \textbf{Sensitivity to hyperparameters}: In the toy models it was tested on, APD only recovers ground-truth mechanisms for a very narrow range of hyperparameters. In particular, APD requires choosing the top-$k$ hyperparameter, the expected number of active parameter components per datapoint, which would usually not be known in advance for non-toy models. As discussed in \citet{braun2025interpretabilityparameterspaceminimizing}, choosing a value for top-$k$ that is too high or low makes it difficult for APD to identify optimal parameter components. 
    \item \textbf{Use of attribution methods}: APD relies on attribution methods (e.g. gradient attributions, used in \citet{braun2025interpretabilityparameterspaceminimizing}), to estimate the causal importance of each parameter component for computing the model's outputs on each datapoint. Gradient-based attributions, and attribution methods more generally, are often poor approximations of ground-truth causal importance \citep{syed2023attributionpatchingoutperformsautomated} and sometimes fail to pass basic sanity checks \citep{adebayo2020sanitycheckssaliencymaps}.

\end{enumerate}

In this work, we introduce a new method for linear parameter decomposition that overcomes all of these issues: \textit{Stochastic Parameter Decomposition} (SPD) (Section \ref{sec:methods}).

Our approach decomposes each matrix in a network into a set of rank-one matrices called \textit{subcomponents}. The number of rank-one matrices can be higher than the rank of the decomposed matrix. Subcomponents are not full parameter components as in the APD method, but they can later be aggregated into full components. In this work, we use toy models with known ground-truth mechanisms, where the clusters are therefore straightforward to identify. However, in future it will be necessary to algorithmically cluster these components in cases where ground truth is not known.

Instead of relying on attribution techniques and a top-$k$ hyperparameter that needs to be chosen in advance, we define the \textit{causal importance} of a subcomponent as how ablatable it is on a given datapoint. Causally important subcomponents should not be ablatable, and ablatable subcomponents should be causally unimportant for computing the output. We train a \textit{causal importance function} to predict the causal importance $g_i \in [0, 1]$ of subcomponent $i$ for computing the model's output on a given datapoint and use the predicted causal importances to ablate unimportant subcomponents by random amounts by masking them multiplicatively with a random scalar sampled from a uniform distribution $\mathcal{U}(g_i, 1)$. We train a model that is parametrized by the sum of these randomly masked subcomponents to compute the same output as the target model.
Crucially, we regularize the predicted causal importance values $g_i$ to be close to zero, so that as many subcomponents as possible will be predicted to be ablatable on any given datapoint. The core intuition behind this training setup is that all combinations of ablating the subcomponents are checked with some probability. And since the causal importance function is penalized for outputting large values of $g$, it should only output high values of $g_i$ when subcomponent $i$ is really ‘used’ by the network.

We apply SPD to all of the toy models that \citet{braun2025interpretabilityparameterspaceminimizing} used to study APD, including: A Toy Model of Superposition \citep{elhage2022toy} (Section \ref{sec:tms}); a Toy Model of Compressed Computation \citep{braun2025interpretabilityparameterspaceminimizing} (Section \ref{sec:residmlp_1layer}); and a Toy Model of Cross-Layer Distributed Representations (Section \ref{sec:residmlp_2layer}). We demonstrate that the method recovers ground-truth mechanisms in all of these models. We also extend the suite of models to include two more challenging models where APD struggles but SPD succeeds: A Toy Model of Superposition with an additional identity matrix in the hidden space (Section \ref{sec:tms_identity}) and a deeper Toy Model of Cross-Layer Distributed Representations (Section \ref{sec:residmlp_2layer}). Using APD, these new models were unmanageably difficult to correctly decompose, but SPD succeeds with relative ease. 

The successful application of SPD to more challenging models demonstrates that SPD is more scalable and stable than APD. Nevertheless, some challenges remain: Firstly, the method needs to be scaled to larger models, which will likely require further improvements in training stability. Second, SPD only finds rank-one components in individual layers, meaning that further clustering step is required to find components that span more than one rank and/or more than one layer. In the toy models presented in this paper, these clusters are known and are therefore straightforward to identify. However, a general clustering solution will be needed in order to find such components where ground-truth is unknown. Despite these challenges, SPD opens up new research avenues for mechanistic interpretability by introducing a linear parameter decomposition method that removes the main barriers to scaling to larger, non-toy models such as language models (Section \ref{sec:discussion}).

\section{Method: Stochastic Parameter Decomposition}\label{sec:methods}
Suppose we have a trained neural network $f(x,W)$ that maps inputs $x$ to outputs $y=f(x,W)$, parametrized by a set of weight matrices\footnote{As in \citet{braun2025interpretabilityparameterspaceminimizing}, we do not decompose biases. Biases can be folded into the weights by treating them as an additional column in each weight matrix, meaning they can in theory be decomposed like any other type of parameter. However, in this work, for simplicity we treat them as their own parameter component that is active for every input, and leave their decomposition for future work.} $W=\{W^1,\dots,W^L\}$. This set $W$ can also be represented as a single vector in a high-dimensional vector space called \textit{parameter space}. Linear parameter decomposition methods such as APD aim to decompose neural network parameters into a set of \textit{parameter components}, which are vectors in parameter space that are trained to exhibit three desirable properties \citep{braun2025interpretabilityparameterspaceminimizing}:

\begin{itemize}
    \item \textbf{Faithfulness:} The parameter components should sum to the parameters of the original network.
    \item \textbf{Minimality:} As few parameter components as possible should be used by the network for a forward pass of any given datapoint in the training dataset. 
    \item \textbf{Simplicity:} Parameter components should use as little computational machinery as possible, in that they should span as few matrices and as few ranks as possible.
\end{itemize}
If a set of parameter components exhibit these three properties, we say that they comprise the network's \textit{mechanisms}\footnote{Note that these properties can trade off against each other. Therefore, in practice, we quantify how much we care about each property, and find the set of parameter components that minimise the resulting overall loss.}. In APD, gradient-based attributions are used to estimate the importance of each parameter component for a given datapoint. Then, the top-$k$ most important parameter components are summed together and used for a second forward pass. These active parameter components are trained to produce the same output on that datapoint as the target model. Simultaneously, the parameter components are trained to sum to the parameters of the target model, and are trained to be simple by penalizing the sum of the spectral $p$-norms of their individual weight matrices, encouraging them to be low-rank. 

In our work, we aim to identify parameter components with the same three properties, but we achieve it in a different way. 

\subsection{Our approach optimizes rank-one subcomponents instead of full-rank parameter components}\label{sec:our-approach-optimizes-rank-one}

A major issue with the APD method is that it is computationally very expensive: It involves optimizing $L$ full-rank matrices for every parameter component (where $L$ is the number of matrices in the model). But this is wasteful if we expect most parameter components to be low-rank and localized only to a subset of layers. With SPD, instead of using full-rank parameter components that span every layer, we decompose each of a neural network's weight matrices $W^1,\dots,W^L$ into a set of $C$ rank-one matrices called \textit{subcomponents}, $\vec{U^l_c} \vec{V_c^{l  \top}}$:
\begin{equation}\label{eq:components}
    W^l_{i,j} \approx \sum^C_{c=1} U^l_{i,c} V^l_{c,j}\,.
\end{equation}
Here, $l$ indexes the neural network's matrices and $i,j$ are its hidden indices.
We will later cluster these subcomponents into full parameter components if they tend to co-activate together. In this paper, the groups are easy to identify and this clustering process is implicit. However, future work will require an explicit clustering algorithm for cases where groups of subcomponents are harder to identify. 

Note that the number of subcomponents in each layer $C$ may be larger than the minimum of the number of rows or columns of the matrix at that layer, thus enabling SPD to identify computations in superposition.

\subsection{Optimizing for faithfulness}\label{sec:optimizing-for-faithfulness}

The way we optimize for faithfulness is the same as in \citet{braun2025interpretabilityparameterspaceminimizing}, by optimizing the sum of our subcomponents to approximate the parameters of the target model:

\begin{equation}
\mathcal{L}_{\text{faithfulness}}=\frac{1}{N}\sum^L_{l=1}\sum_{i,j}{\left( W^{l}_{i,j}- \sum^C_{c=1} U^l_{i,c} V^l_{c,j}\right)}^2 ,
\end{equation}
where $N$ is the total number of parameters in the target model.
\subsection{Optimizing for minimality and simplicity by learning a causal importance function to stochastically sample masks} \label{sec:optimizing-for-minimality-and-simplicity}

The way we optimize for minimality and simplicity is different from \citet{braun2025interpretabilityparameterspaceminimizing}. Since we already start with rank-one subcomponents that are localized in single layers, we don't need to optimize for subcomponent simplicity. 
Instead, we only need to train our set of subcomponents such that as few as possible are "\textit{active}" or "\textit{used}" or "\textit{required}" by the network to compute its output on any given datapoint in the training set. We consider this equivalent to requiring that as few subcomponents as possible be \textit{causally important} for computing the network's output. 

To optimize our set of subcomponents such that as few as possible are causally important for computing the model's output on any given datapoint, we have three requirements:
\begin{enumerate}
    \item A formal definition of what it means for a subcomponent to be `causally important' for computing the model's outputs (Section \ref{sec:defining-causal-importance});
    \item A loss function that trains causally important subcomponents to compute the same function as the original network (Section \ref{sec:loss-function-optimize-subcomponents});
    \item A loss function that encourages as many subcomponents as possible to be causally \textit{un}-important on each datapoint (Section \ref{sec:loss-function-causally-unimportant}).
\end{enumerate}

\subsubsection{Requirement 1: Formally defining a subcomponent's \textit{causal importance} as the extent to which it can be ablated}\label{sec:defining-causal-importance}

Intuitively, we say a subcomponent is \textit{causally important} on a particular datapoint $x$ if and only if it is required to compute the model's output for that datapoint. Conversely, we say a subcomponent is \textit{causally unimportant} if it can be ablated by some arbitrary amount while leaving the model's output unchanged. In particular, if a component is fully unimportant, then it shouldn't matter how much we ablate it by; we should be able to fully ablate all causally unimportant components, or only partially ablate them, or ablate only a subset of them, and still get the same model output. Note, the distinction between causally important and unimportant is not binary; we can say that a subcomponent is causally unimportant \textit{to the extent} that it can be ablated without affecting the model's output. 

Formally, suppose $g^l_c(x)\in[0,1]$ indicates the causal importance of subcomponent $(c)$ of weight matrix $l$ on a given datapoint $x$. For now, we take this quantity as given, but later we discuss how we obtain it. In general, we want to get the same model output for any weight matrices along all monotonic `ablation curves' $m^l_c(x,r)$ that interpolate between the original model and the model with all inactive subcomponents ablated. Here, $r$ is a vector sampled such that:
\begin{equation}\label{eq:subcomponents}
\begin{aligned}
&r^l_c \in [0, 1]\\
&m^l_c(x,r):=g^l_c(x)+(1-g^l_c(x))r^l_c\\ 
&W'^l_{i,j}(x,r):=\sum^C_{c=1} U^l_{i,c} m^l_c(x,r) V^l_{c,j}  \\
&\forall r: f(x\vert W'^1(x,r),\dots,W'^L(x, r))\approx f(x\vert W^1,\dots,W^L)\,.
\end{aligned}
\end{equation}
Now we need a differentiable loss function(s) in order to be able to train the masked model $f(x\vert W'^1(x,r),\dots,W'^L(x, r))$ to approximate the target model $f(x\vert W^1,\dots,W^L)$ for all values of $r\in {[0,1]}^{C\times L}$. 

\subsubsection{Requirement 2: A loss function that lets us optimize causally important subcomponents to approximate the same function as the original network} \label{sec:loss-function-optimize-subcomponents}

Unfortunately, calculating $f(x\vert W'^1(x,r),\dots,W'^L(x, r))$ for all values of $r\in {[0,1]}^{C\times L}$ is computationally intractable because there are infinitely many possible values. But we can calculate it for a randomly chosen subset of values. To do this, we uniformly sample $S$ points, $r^{l,(s)}_{c} \sim \mathcal{U}(0,1)$ and mask the subcomponents with the resulting stochastic masks $m^l_c(x, r^{(s)})$. Then, as $S\rightarrow \infty$, the following loss approximately optimizes $f(x\vert W'^1(x,r),\dots,W'^L(x, r))$ to satisfy Equation $\ref{eq:subcomponents}$:
\begin{equation}\label{eq:random_recon}
\begin{aligned}
\mathcal{L}_{\text{stochastic-recon}}&=\frac{1}{S}\sum^S_{s=1}D \left( f(x\vert W'(x,r^{(s)})),f(x\vert W) \right)\,\\
\end{aligned}
\end{equation}
Here, $D$ is some appropriate divergence measure in the space of model outputs, such as KL-divergence for language models, or MSE loss.

However, this loss can be somewhat noisy because it involves a forward pass in which every subcomponent has been multiplied by a random mask. Therefore, in addition to this loss, we also use an auxiliary loss $\mathcal{L}_{\text{stochastic-recon-layerwise}}$, which is simply a layerwise version of $\mathcal{L}_{\text{stochastic-recon}}$ where only the parameters in a single layer at a time are replaced by stochastically masked subcomponents. The gradients are still calculated at the output of the model: 

\begin{equation}\label{eq:layerwise_random_recon}
\begin{aligned}
\mathcal{L}_{\text{stochastic-recon-layerwise}}=\frac{1}{LS}\sum^L_{l=1}\sum^S_{s=1}D \left( f(x\vert W^1,\dots,W'^l(x,r^{l, (s)}),\dots,W^L),f(x\vert W) \right)\,
\end{aligned}
\end{equation}
This should not substantially alter the global optimum of training, because Equation \ref{eq:layerwise_random_recon} is equivalent to Equation \ref{eq:random_recon} if the subcomponents sum to the original weights and if we sample $r^l_c=1$ for all layers except one at a time.  

\subsubsection{Requirement 3: A loss function that encourages as many subcomponents as possible to be causally unimportant} \label{sec:loss-function-causally-unimportant}

We have not yet defined how we obtain subcomponents' causal importance values $g^l_c(x)$. Measuring this directly would be intractable, so we learn a function to predict it. 

In theory, we could use any arbitrary \textit{causal importance function} $\Gamma: X \rightarrow [0,1]^{C \times L}$ to predict causal importance values. In practice, we use a set of independent functions $\Gamma^l_c$, one for each subcomponent. Each function consists of a very small trained MLP $\gamma^l_c$ (see Appendix \ref{sec:architecture-causal-importance-mlps} for its architecture). Each MLP takes as input a scalar, the subcomponent's `inner activation', $h^l_c(x):=\sum_j V^l_{c,j} a^l_j(x)$, where $\vec{a^l(x)}$ is the activation vector in the target model that gets multiplied by weight matrix $W^l$. The MLP outputs are passed to a hard sigmoid $\sigma_H$ to get a prediction of the subcomponent's causal importance\footnote{We use hard sigmoids rather than standard sigmoids because we would like to make it possible for causal importance values to take values of exactly $0$ or $1$ or anywhere in between; a standard sigmoid function only tends toward $0$ or $1$ as its input tends toward $-\infty$ or $\infty$ respectively. However, note that hard sigmoids have large regions where gradients are $0$. We therefore actually use leaky-hard sigmoids to increase training stability.
Outputs may therefore lie slightly outside of the $[0,1]$ range (Appendix \ref{sec:avoiding-dead-gradients}).}:
\begin{equation}
    g^l_c(x)= \Gamma^l_c (x) = \sigma_H ( \gamma^l_c ( h^l_c(x) ) )
\end{equation}

The output of the causal importance function for each subcomponent is therefore a single scalar number that should be in the range $[0,1]$. 

While this form of causal importance function works well for the toy models in this paper, it is likely that the best causal importance functions for arbitrary models require more expressivity. For example, a subcomponent's causal importance function may take as input the inner activations of all subcomponents, rather than just its own. Such variants may be explored in future work.

The causal importance values $g^l_c(x)$ are used to sample masks that randomly ablate each subcomponent $m^l_c(x, g^l_c(x)) \sim \mathcal{U}(g^l_c(x), 1)$.
We use the reparametrization trick \citep{kingma2013autoencodingvariationalbayes}(Equation \ref{eq:subcomponents}) to allow gradients to be backpropagated through the masks $m^l_c(x, g^l_c(x))$ to train the causal importance functions $\Gamma^l_c$.
The causal importance functions can therefore learn to produce masks that better predict the ablatability of each subcomponent on a given datapoint. However, if the causal importance functions were trained using only the $\mathcal{L}_{\text{stochastic-recon}}$ loss, they could perform optimally (but pathologically) by outputting a causal importance value of $g^l_c(x)=1$ for every subcomponent on every input, since there is no incentive to learn to predict the full extent of the ablatability of the subcomponents. An additional loss is required to encourage the causal importance function to always try to ablate as much as possible. We therefore penalize causal importance values for being above $0$ using a $\mathcal{L}_{\text{importance-minimality}}$ loss:
\begin{equation}\label{eq:minimal}
\begin{aligned}
\mathcal{L}_{\text{importance-minimality}}=\sum^L_{l=1}\sum^C_{c=1} \vert g^l_c(x) \vert^p\,,
\end{aligned}
\end{equation}
where $p>0$ \footnote{It is important to note that $\mathcal{L}_{\text{importance-minimality}}$ is somewhat different from the $L_p$ penalties often used to sparsify latents in SDL, where $p$ is restricted to $(0,1]$. We find that training with $\mathcal{L}_{\text{importance-minimality}}$ successfully minimizes importance values even when $p>1$. We believe that this happens for the following reason: If $p>1$ in SDL, a single active feature can always be split into many active features to improve the sparsity loss. However, if $p>1$ in SPD, then if one causally important subcomponent were pathologically split into two, both resulting subcomponents would still have causal importance of $\approx 1$, since both are still needed to maintain low $\mathcal{L}_{\text{stochastic-recon}}$, meaning $\mathcal{L}_{\text{importance-minimality}}$ would increase.}.

\subsection{Summary of training setup}\label{sec:method_summary}

Our full loss function consists of four losses: 

\begin{equation}\label{eq:full_loss}
\mathcal{L}_{\text{SPD}} = \mathcal{L}_{\text{faithfulness}}+ ( \beta_1 \mathcal{L}_{\text{stochastic-recon}}+ \beta_2 \mathcal{L}_{\text{stochastic-recon-layerwise}} ) + \beta_3 \mathcal{L}_{\text{importance-minimality}}
\end{equation}

Our training setup involves five hyperparameters (excluding optimizer hyperparameters such as learning rate): The coefficients $\beta_1, \beta_2, \beta_3$ for each of the losses; the $p$-norm used in $\mathcal{L}_{\text{importance-minimality}}$ and the number of mask samples $S$ that are used for each training step, for which we find $S=1$ sufficient. We discuss some heuristics for choosing the loss coefficients in Appendix \ref{sec:hyperparam-heuristics}. Pseudocode for the SPD algorithm can be found in Appendix \ref{app:pseudocode}.

\section{Results}\label{sec:results}
We apply SPD to decompose a set of toy models with known ground-truth mechanisms. Some of these models were previously studied by \citet{braun2025interpretabilityparameterspaceminimizing} to evaluate APD, while others are new. We study:
\begin{enumerate}

    \item A \textbf{Toy Model of Superposition (TMS)} \citep{elhage2022toy} previously studied by \citet{braun2025interpretabilityparameterspaceminimizing} (Section \ref{sec:tms});
    \item A \textbf{TMS model with an identity matrix} inserted in the middle of the model (not previously studied) (Section \ref{sec:tms_identity});
    \item A \textbf{Toy Model of Compressed Computation} previously studied by \citet{braun2025interpretabilityparameterspaceminimizing} (Section \ref{sec:residmlp_1layer}); 
    \item Two \textbf{Toy Models of Cross-Layer Distributed Representations}: One with two residual MLP blocks (previously studied by \citet{braun2025interpretabilityparameterspaceminimizing}) and one with three MLP blocks (not previously studied) (Section \ref{sec:residmlp_2layer});
\end{enumerate}
In all cases, we find that SPD seems to identify known ground-truth mechanisms up to a small error. For the models that were also decomposed with APD in \citet{braun2025interpretabilityparameterspaceminimizing}, we find that the SPD decompositions have fewer errors despite requiring less hyperparameter tuning to find the ground-truth mechanisms.

Code to reproduce our experiments can be found at \url{https://github.com/goodfire-ai/spd/tree/spd-paper}.
Training details and hyperparameters can be found in Appendix \ref{app:training-details}. Additional figures and training logs can be found in the WandB report \href{https://wandb.ai/goodfire/spd-tms/reports/SPD-paper-report--VmlldzoxMzE3NzU0MQ?accessToken=427spmsbxig5cyp4jsprg9p183tysclk7ttzyxjlsiwafh8badzlpgxcvopsormm}{here}.

\begin{figure}
    \centering
    \includegraphics[width=1\linewidth]{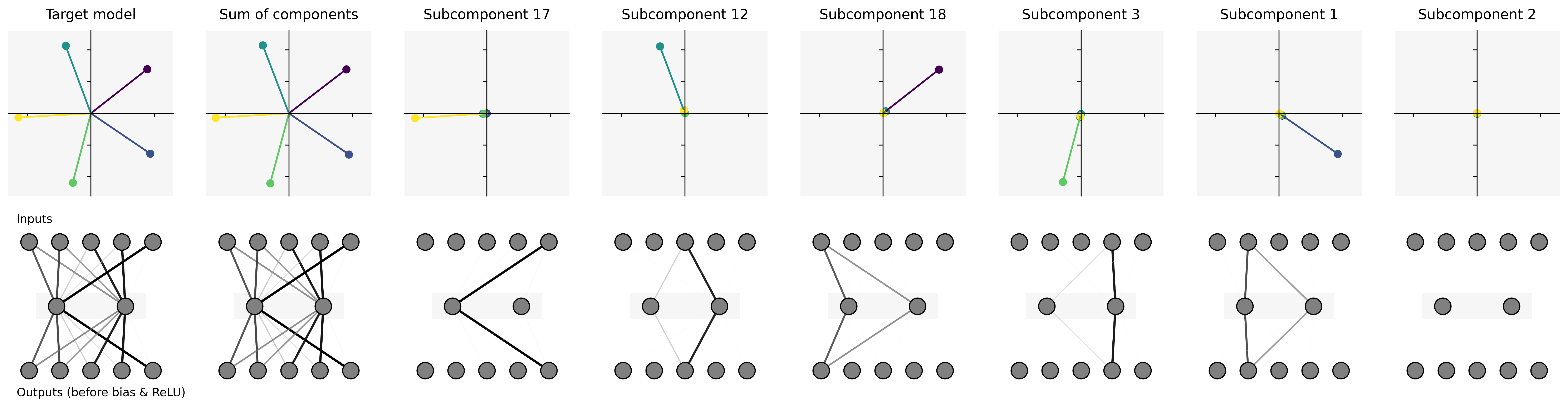}
    \caption{Results of running SPD on $\text{TMS}_{5-2}$. \textbf{Top row:} Plots of (left to right) the columns of the weight matrix of the target model; the sum of the SPD parameter components; and individual parameter components. Although this run of SPD used 20 subcomponents, only 6 subcomponents are shown, ordered by the sum of the norms of each of the columns of their (rank-one) weight matrices. The first five have learned one direction each, each corresponding to one of the columns of the target model. The final column and the other 14 components (not shown) have a negligible norm because they are superfluous for replicating the behavior of the target model. \textbf{Bottom row:} Depiction of the corresponding parametrized networks.}
    \label{fig:tms-combined}
\end{figure}

\begin{table}[t]
\centering
\begin{tabular}{lccc}
\toprule
& MMCS & ML2R \\
\midrule
$\text{TMS}_{5-2}$ & 1.000 ± 0.000 & 0.993 ± 0.002  \\
$\text{TMS}_{40-10}$ & 1.000 ± 0.000 & 1.010 ± 0.007  \\
\midrule
$\text{TMS}_{5-2+\text{ID}}$ & 1.000 ± 0.000 & 0.992 ± 0.010  \\
$\text{TMS}_{40-10+\text{ID}}$ & 1.000 ± 0.000 & 1.031 ± 0.001  \\
\bottomrule
\end{tabular}
\vskip 0.15in
\caption{Mean Max Cosine Similarity (MMCS) and Mean L2 Ratio (ML2R) with their standard deviations (to $3$ decimal places) between learned parameter subcomponents and the target model weights in the subcomponents found by SPD for the embedding matrix $W$ matrix in the $\text{TMS}_{5-2}$ and $\text{TMS}_{40-10}$ models (Section \ref{sec:tms}) and $\text{TMS}_{5-2+\text{ID}}$ and $\text{TMS}_{40-10+\text{ID}}$ models (Section \ref{sec:tms_identity}). These results indicate that the ground truth mechanisms are recovered perfectly and with negligible shrinkage for all models.}
\label{tab:tms_mmcs}
\end{table}

\subsection{Toy Model of Superposition}\label{sec:tms}
We decompose \citet{elhage2022toy}'s Toy Model of Superposition (TMS), which can be written as $\hat{x}= \text{ReLU}(W^\top W x + b)$, with weight matrix $W \in \mathbb{R}^{m_1 \times m_2}$.
The model is trained to reconstruct its inputs, which are sparse sums of one-hot $m_2$-dimensional input features whose activations are scaled to a random uniform distribution $[0,1]$. 
Typically, $m_1 < m_2$, so the model is forced to `squeeze' representations through a $m_1$-dimensional bottleneck. When the model is trained on sufficiently sparse data distributions, it can learn to represent features in superposition in this bottleneck. For certain values of $m_1$ and $m_2$, the columns of the W matrix can form regular polygons in the $m_1$-dimensional hidden activation space (Figure \ref{fig:tms-combined} - Leftmost panel).

The ground truth mechanisms in this model should be a set of rank-$1$ matrices that are zero everywhere except in the $c^{\text{th}}$ column, where they take the values $\vec{W_{:, c}}$. Intuitively, a column of $W$ is only `used' if the corresponding input feature is active; other columns can be ablated without significantly affecting the output of the model.

\subsubsection*{SPD Results: Toy Model of Superposition}

We apply SPD to TMS models with $5$ features in $2$ dimensions, denoted $\text{TMS}_{5-2}$ and $40$ features in $10$ dimensions, denoted $\text{TMS}_{40-10}$. In both cases, SPD successfully decomposes the model into subcomponents that closely correspond to the columns of $W$ (Figure \ref{fig:tms-combined}). This result is robust to different training seeds and required less hyperparameter tuning than APD \citep{braun2025interpretabilityparameterspaceminimizing}. 

We quantify how aligned the learned parameter components vectors are to the columns of $W$ in the target model using the mean max cosine similarity (MMCS) \citep{Sharkey_Braun_Millidge_2022}. The MMCS measures alignment between each column of $W$ and the corresponding column in the parameter component that it best aligns with:
\begin{equation}\label{eq:mmcs}
\text{MMCS}(W, \{\vec{U_c} \vec{V_c}^\top\}) = \frac{1}{m_2}\sum_{j=1}^{m_2}\max_c \left( \frac{\vec{U_{:,c}} V_{c,j} \cdot W_{:,j}}{\vert\vert \vec{U_{:,c}} V_{c,j} \vert\vert_2 \vert\vert W_{:,j}\vert\vert_2} \right) \,,
\end{equation}

where $c\in C$ are parameter component indices and $j\in \{ 1, \cdots, m_2\}$ are input feature indices. 
A value of $1$ for MMCS indicates that, for all input feature directions in the target model, there exists a parameter subcomponent whose corresponding column points in the same direction. 

We also quantify how close their magnitudes are with the mean L2 Ratio (ML2R) between the Euclidean norm of the columns of $W$ and the Euclidean norm of the columns of the parameter components with which they have the highest cosine similarity:
\begin{equation}\label{eq:ml2r}
\text{ML2R}(W, \{\vec{U_c} \vec{V_c}^\top\}) = \frac{1}{m_2}\sum_{j=1}^{m_2} \frac{\vert\vert \vec{U}_{:,\text{mcs}(j)} V_{\text{mcs}(j),j}\vert\vert_2}{\vert\vert W_{:,j}\vert\vert_2}\,,
\end{equation}
where $\text{mcs}(j)$ is the index of the subcomponent that has maximum cosine similarity with weight column $j$ of the target model.
A value close to $1$ for the ML2R indicates that the magnitude of each parameter component is close to that of its corresponding target model column.

For both $\text{TMS}_{5-2}$ and $\text{TMS}_{40-10}$, the MMCS and ML2R values are $\approx 1$, with a closer match than that obtained for the decomposition in \citep{braun2025interpretabilityparameterspaceminimizing} (Table \ref{tab:tms_mmcs}). This indicates that the parameter components are close representations of the target model geometrically. In contrast, APD exhibited significant shrinkage of the learned parameter components, with an ML2R of approximately 0.9 for both models \citep{braun2025interpretabilityparameterspaceminimizing}, reminiscent of feature shrinkage in SAEs \citep{jermyn2024tanh, wright2024suppression}.

\begin{figure}
    \centering
    \includegraphics[width=1\linewidth]{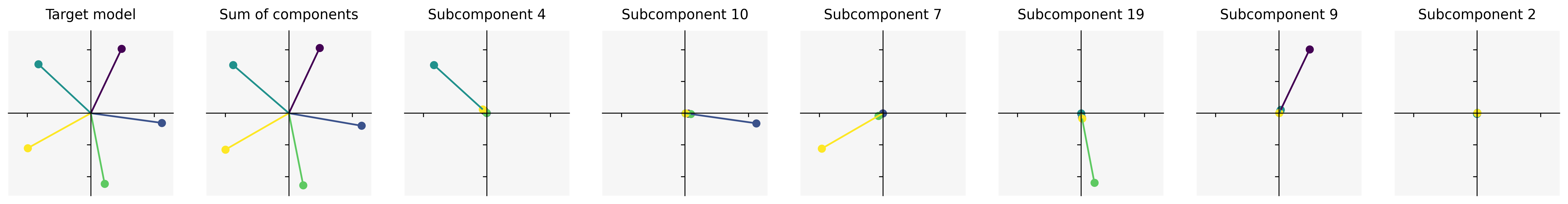}
    \caption{Plots of (left to right) the columns of the input weight matrix $W$ of the $\text{TMS}_{5-2+\text{ID}}$ model (which can be written as $\hat{x}= \text{ReLU}(W^\top I W x + b)$, with a weight matrix $W \in \mathbb{R}^{m_1 \times m_2}$ and an identity matrix $I\in \mathbb{R}^{m_1 \times m_1}$); the sum of the parameter subcomponents for that matrix found by SPD; and the individual parameter subcomponents. Although this run of SPD used $20$ subcomponents, only $6$ subcomponents are shown, ordered by the sum of their matrix norms. The first five have learned one direction each, each corresponding to one of the columns of the target model. The final column and the other 14 components (not shown) have a negligible norm because they are superfluous for replicating the behavior of the target model.}
    \label{fig:tms-identity-input-weights}
\end{figure}

\begin{figure}
    \centering
    \includegraphics[width=0.8\linewidth]{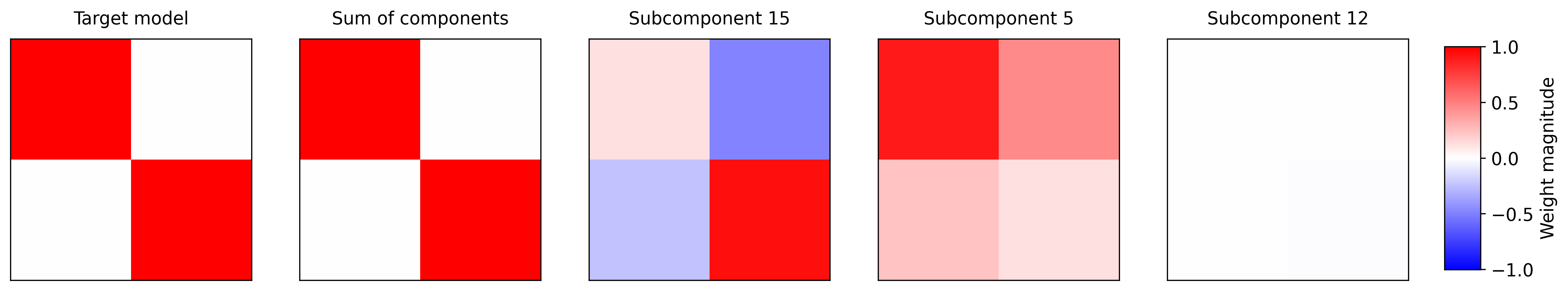}
    \caption{Plots of (left to right) the weights in the hidden identity matrix $I$ of the $\text{TMS}_{5-2+\text{ID}}$; the sum of all subcomponents found by SPD for that matrix (including small-norm subcomponents that are not shown); and the largest three individual subcomponents. We see that SPD finds two subcomponents that together sum to the original rank-$2$ identity matrix of the target model, while the other subcomponents have a negligible weight norm.}
    \label{fig:tms-identity-hidden-weights}
\end{figure}

\begin{figure}
    \centering
    \includegraphics[width=1\linewidth]{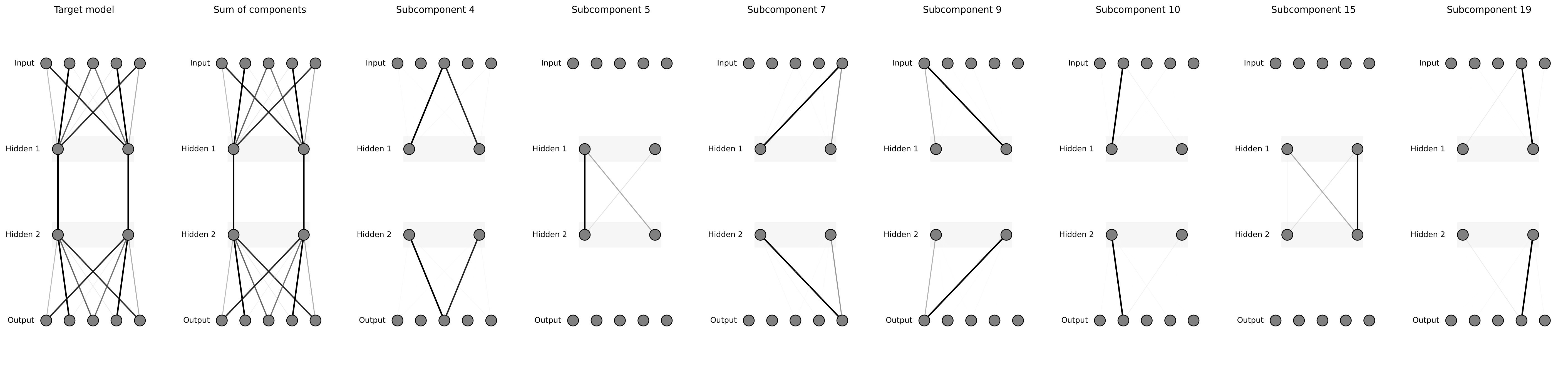}
    \caption{Plots of (left to right) the $\text{TMS}_{5-2+\text{ID}}$ networks parametrized by: The target model parameters; the sum of all parameter subcomponents found by SPD the decomposition of the model; and the seven individual subcomponents of non-negligible size. We see that SPD finds five subcomponents for the embedding matrix $W$, corresponding to the five input features, and two subcomponents that span the identity matrix $I$ in the middle of the model.}
    \label{fig:tms-identity-full-network}
\end{figure}

\subsection{Toy Model of Superposition with hidden identity}\label{sec:tms_identity}

SDL methods are known to suffer from the phenomenon of `feature splitting', where the features that are learned depend on the dictionary size, with larger dictionaries finding more sparsely activating, finer-grained features than smaller dictionaries \citep{bricken2023monosemanticity, chanin2024absorptionstudyingfeaturesplitting}: Suppose a network has a hidden layer that simply implements a linear map. When decomposing this layer with a transcoder, we can continually increase the number of latents and learn ever more sparsely activating, ever more fine-grained latents to better minimize its reconstruction and sparsity losses. This problem is particularly salient in the case of a linear map, but similar arguments apply to nonlinear maps. 

By contrast, \citet{braun2025interpretabilityparameterspaceminimizing} claimed that linear parameter decomposition methods do not suffer from feature splitting. In the linear case, SPD losses would be minimized by learning a
single $d$-dimensional component that performs the linear map. The losses cannot be further reduced by adding more subcomponents, because that would prevent the components from summing to the original network weights.

Here, we empirically demonstrate this claim in a simple setting. We train a toy model of superposition identical to the $\text{TMS}_{5-2}$ and $\text{TMS}_{40-10}$, but with identity matrices inserted between the down-projection and up-projection steps of the models. These models, denoted $\text{TMS}_{5-2+\text{ID}}$ and $\text{TMS}_{40-10+\text{ID}}$, can be written as $\hat{x}= \text{ReLU}(W^\top I W x + b)$, with a weight matrix $W \in \mathbb{R}^{m_1 \times m_2}$ and an identity matrix $I\in \mathbb{R}^{m_1 \times m_1}$.

We should expect SPD to find $m_2 + m_1$ subcomponents in total: $m_2$ subcomponents that correspond to the columns of the matrix $W$ (each having causal importance for the model output if and only if one particular feature is present in the input, as in standard TMS (Section \ref{sec:tms})) and $m_1$ subcomponents for the matrix $I$ (almost all of which should have causal importance for the model output on every input). There are $m_1$ subcomponents because we need $m_1$ rank-one matrices to sum to a rank-$m_1$ matrix.

\subsubsection*{SPD Results: Toy Model of Superposition with hidden identity}

As expected, we find that SPD decomposes the embedding matrices $W$ of both the $\text{TMS}_{5-2+\text{ID}}$ and $\text{TMS}_{40-10+\text{ID}}$ models into parameter subcomponents that closely correspond to their columns (Figure \ref{fig:tms-identity-hidden-weights}; Table \ref{tab:tms_mmcs}).

Also as expected, SPD decomposes the identity matrix $I\in \mathbb{R}^{m_1 \times m_1}$ into only $m_1$ subcomponents that sum to the original matrix $I$ (Figure \ref{fig:tms-identity-hidden-weights}). In total, SPD identifies only $m_2 + m_1$ subcomponents with non-negligible norm, thus identifying the ground truth mechanisms in each layer (Figure \ref{fig:tms-identity-full-network}). APD fails to learn ground truth mechanisms in this model.

\subsection{Toy Model of Compressed Computation} \label{sec:residmlp_1layer}

\begin{figure}
    \centering
    \includegraphics[width=0.4\linewidth]{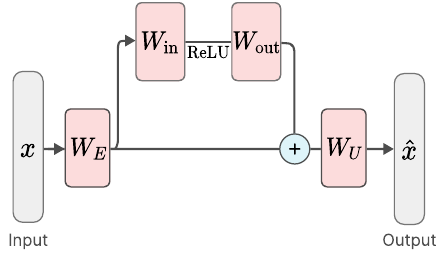}
    \caption{The architecture of the Toy Model of Compressed Computation. It uses a $1$-layer residual MLP. Figure adapted from \citet{braun2025interpretabilityparameterspaceminimizing}.}
    \label{fig:resid_mlp_1l_architecture}
\end{figure}

In this setting, the target network is a residual MLP that was previously studied by \citet{braun2025interpretabilityparameterspaceminimizing} (Figure \ref{fig:resid_mlp_1l_architecture}). It consists of a single residual MLP layer of width $d_{\rm mlp}=50$; a fixed, random embedding matrix with unit norm rows $W_E$; an unembedding matrix $W_U=W_E^\top$; and $100$ input features, with a residual stream width of $d_{\rm resid}=1000$. This model is trained to approximate a function of sparsely activating input features $x_i\in[-1, 1]$, using a Mean Squared Error (MSE) loss between the model output and the labels. The labels we train the model to predict are produced by the function $y_i = x_i + \text{ReLU}(x_i)$. Crucially, the task involves learning to compute more ReLU functions than the network has neurons.

A naive solution to this task would be to dedicate each of the $50$ neurons `monosemantically' to computing one of the $100$ input-output mappings. But this solution would perform poorly on the other $50$ mappings. Instead, the model seems to achieve a better loss by using multiple neurons `polysemantically' to compute a single input-output mapping. We thus expected that, for each mapping, a single subcomponent in the model's MLP input weight matrix $W_\text{in}$ would be responsible for computing it, with every component connecting to many neurons\footnote{See \citet{BhagatCC} for further discussion on how the model may be performing this task.}. 

Originally, we were unsure what the ground-truth mechanisms in the model's MLP output weight matrix $W_\text{out}$ were. It seemed both possible that (a) for each input-output mapping, the model would have a single mechanism within $W_\text{out}$ that projects the result of MLP activations back into the residual stream, or (b) that $W_\text{out}$ was a single mechanism embedding the entire MLP output back into the residual stream, similar to the case of the identity matrix in the TMS model in Section \ref{sec:tms_identity}. The analysis based on the APD decomposition in \citet{braun2025interpretabilityparameterspaceminimizing} claimed the former, 
but our SPD decomposition, as shown below, clearly finds the latter. 

\subsubsection*{SPD Results: Toy Model of Compressed Computation}

To understand how each neuron participates in computing the output for a given input feature, \citet{braun2025interpretabilityparameterspaceminimizing} measured the neuron's \textit{contribution} to each input feature computation.
The neuron contributions for input features $i\in \{ 0, \cdots, 99 \}$ are calculated as:
\begin{equation}\label{eq:neuron-contribution-model}
    \text{NeuronContribution}_{\text{model}}(i) = ({W_U}_{[i,:]} W_\text{out}) \odot (W_\text{in} {W_E}_{[:,i]})
\end{equation}
where ${W_E}_{[:,i]}, {W_U}_{[i,:]}$ are the $i$-th column of the embedding matrix and the $i$-th row of the unembedding matrix, and $\odot$ denotes element-wise multiplication. A large positive contribution indicates that the neuron plays an important role in computing the output for input feature $i$. 

Neuron contributions for individual subcomponents of $W_{\text{in}}$ are calculated in a similar way:
\begin{equation}\label{eq:neuron-contribution-subcomponents}
    \text{NeuronContribution}_{\text{subcomponent}}(i) = \max_c [({W_U}_{[i,:]} U_\text{out} V_\text{out}^\top) \odot (U_{\text{in}, [:,c]} V_{\text{in}, [c,:]}^\top {W_E}_{[:,i]})]
\end{equation}

\begin{figure}
    \centering
    \includegraphics[width=0.85\linewidth]{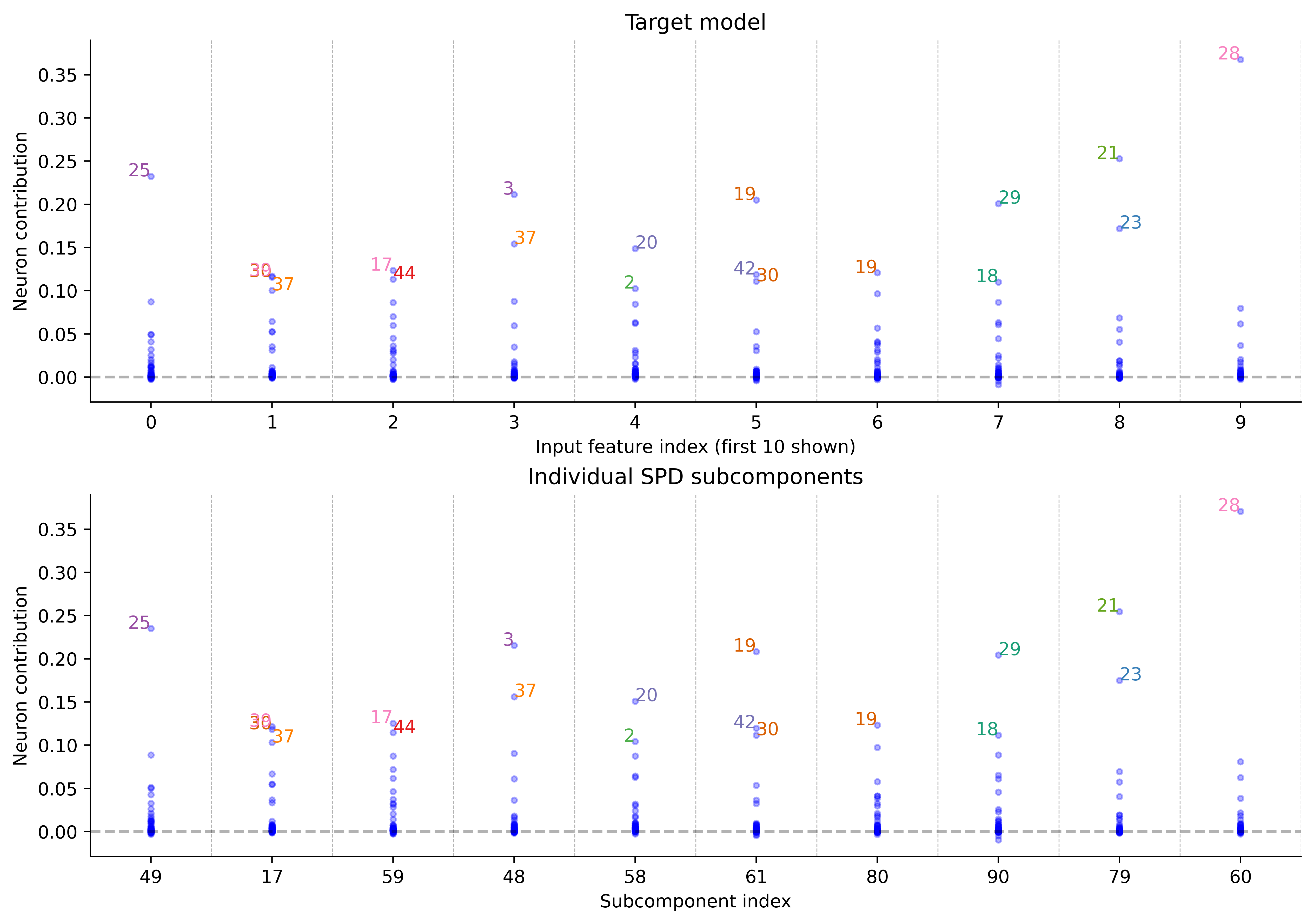}
    \caption{Toy Model of Compressed Computation: Similarity between target model weights and SPD subcomponents for the first $10$ (out of $100$) input feature dimensions. \textbf{Top}: Neuron contributions measured by Equation \ref{eq:neuron-contribution-model} for each input feature index $i\in \{0,\dots,9\}$.  \textbf{Bottom}: Neuron contributions for the corresponding parameter subcomponents, measured by Equation \ref{eq:neuron-contribution-subcomponents} for each input feature index $i\in \{0,\dots,9\}$. The neurons are numbered from $0$ to $49$ based on their raw position in the MLP layer. An extended version of this figure showing all input features and parameter components can be found \href{https://wandb.ai/goodfire/spd-tms/reports/SPD-paper-report--VmlldzoxMzE3NzU0MQ?accessToken=427spmsbxig5cyp4jsprg9p183tysclk7ttzyxjlsiwafh8badzlpgxcvopsormm}{here}.}
    \label{fig:resid-mlp-weights}
\end{figure}

\begin{figure}
    \centering
    \includegraphics[width=0.5\linewidth]{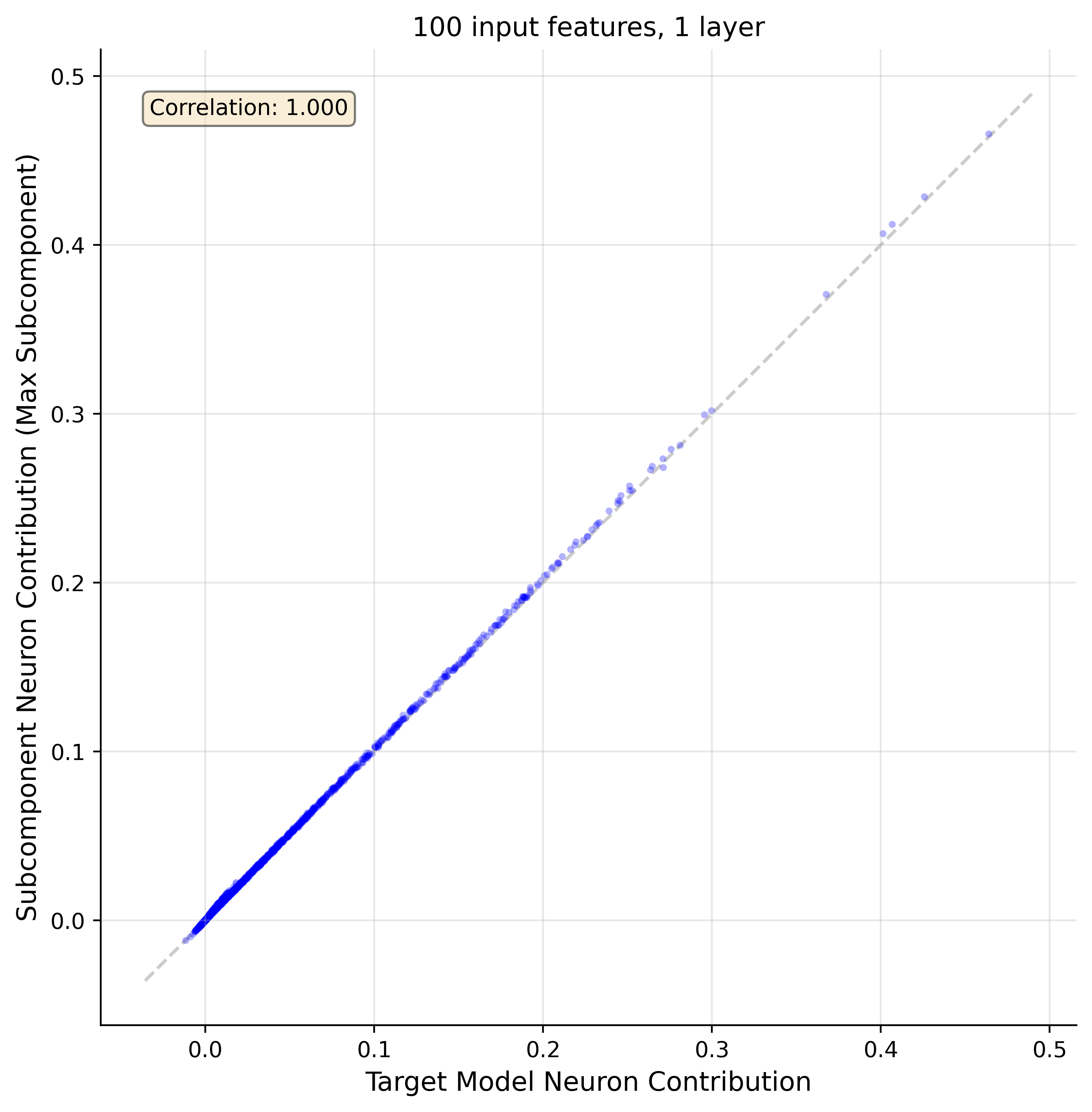}
    \caption{Toy Model of Compressed Computation: Similarity between target model weights and SPD subcomponents for all $100$ input feature dimensions. \textbf{X-axis}: Neuron contributions measured by Equation \ref{eq:neuron-contribution-model} for each input feature index $i\in \{0,\dots,99\}$.  \textbf{Y-axis}: Neuron contributions for the corresponding parameter subcomponents of $W_{\text{in}}$, measured by Equation \ref{eq:neuron-contribution-subcomponents} for each feature index $i\in \{0,\dots,99\}$. There is a very close match between the X and Y axis for each neuron contribution, indicating that each subcomponent connects its corresponding feature to the MLP neurons with almost the same weights as the target model.}
    \label{fig:resid-mlp-weights-all}
\end{figure}

\begin{figure}
    \centering
    \includegraphics[width=1\linewidth]{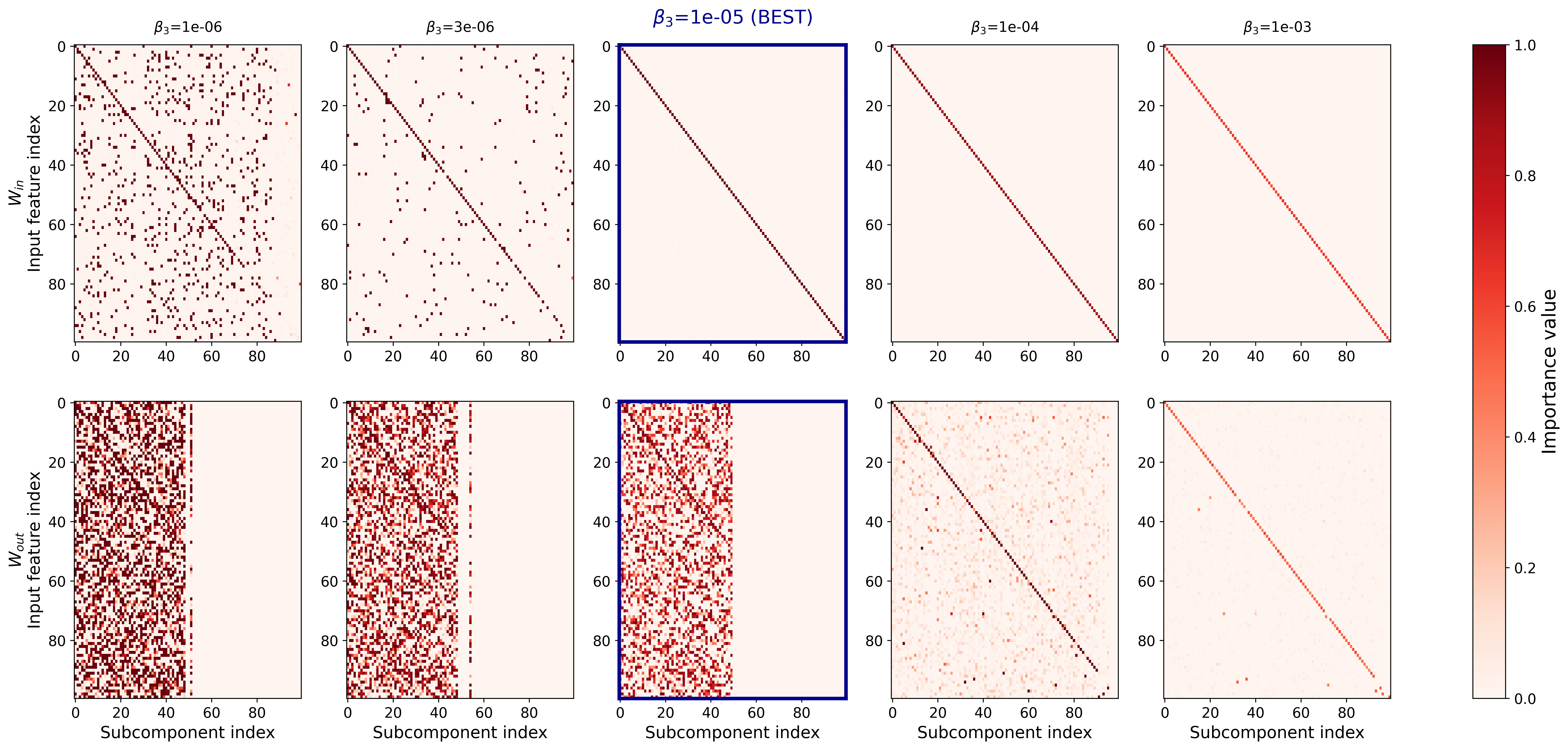}
    \caption{Causal importance values of each subcomponent (clipped between $0$ and $1$) in response to one-hot inputs ($x_i=0.75$) for multiple importance loss coefficients $\beta_3$. The columns of each matrix are permuted differently, ordered by iteratively choosing the subcomponent (without replacement) with the highest causal importance for each input feature. When the importance loss coefficient $\beta_3$ is too low (Left), subcomponents in $W_{\text{in}}$ are not `monosemantic' (i.e. multiple subcomponents have causal importance for the same feature). When the importance loss coefficient $\beta_3$ is just right (middle column), each subcomponent in $W_{\text{in}}$ has causal importance for computing a unique input feature. It also identifies the correct number of subcomponents in $W_{\text{out}}$ ($50$). Although it identifies the correct number of components, these subcomponents need not align with any particular basis, and hence look `noisy' because they align with multiple features. But this does not matter, since they always co-activate together and sum to the target model's identity-matrix parameters. When the importance loss coefficient $\beta_3$ is too high (Right), the rank-$50$ $W_{\text{out}}$ component is split into too many subcomponents---approximately one subcomponent for each feature, but where many of the subcomponents have small causal importance values for other features. Also, the causal importance values on the diagonal shrink far below $1.0$, resulting in high $\mathcal{L}_{\text{stochastic-recon}}$ and $\mathcal{L}_{\text{stochastic-recon-layerwise}}$ losses.}
    \label{fig:resid-mlp-varying-sparsity-importance-vals}
\end{figure}

We apply SPD to the MLP weight matrices $W_\text{in}, W_\text{out}$ in the target model. We find that SPD decomposes the $W_\text{in}$ matrix into $100$ subcomponents, each of which is involved in implementing $y_i = x_i + \text{ReLU}(x_i)$ for a unique input dimension $i\in \{0,\cdots,99 \}$. The computations required to produce the output for each input dimension in the target network (Figure \ref{fig:resid-mlp-weights} - top) are well replicated by individual parameter subcomponents in the SPD model (Figure \ref{fig:resid-mlp-weights} - bottom). For each input dimension, there is a corresponding parameter subcomponent of $W_\text{in}$ that uses the same neurons to compute the function as the target model does. In other words, for each input dimension, the neuron contributions for each input dimension match the neuron contributions of some corresponding subcomponent. Figure \ref{fig:resid-mlp-weights-all} also summarizes all neuron contributions in the target model compared to individual subcomponents in the SPD model and shows they all match closely. Given a one-hot input $x_i$ of magnitude $0.75$ to the model, the masking functions in $W_\text{in}$ select only their corresponding component to activate (Figure \ref{fig:resid-mlp-varying-sparsity-importance-vals} - middle column).

Meanwhile, SPD splits its corresponding $W_\text{out}$ into $50$ subcomponents, which appear to all be effectively part of a single rank-$50$ component comprising the entire $W_\text{out}$ matrix, since the masking functions seem to activate many components despite being a one-hot input $x_i=0.75$ to the model. Both APD and SPD split the MLP input weight matrices $W_\text{in}$ into one component per input feature \citep{braun2025interpretabilityparameterspaceminimizing}. However, APD also decomposed the MLP output weight matrix $W_\text{out}$ into $100$ components roughly corresponding to different input features, instead of a single rank-$50$ component. We believe that the SPD decomposition is more accurate, and that the APD decomposition of $W_\text{out}$ was incorrect due to a poorly chosen value for the top-$k$ hyperparameter. The SPD decomposition seems to reconstruct the computations of the original model more cleanly. In particular, \citet{braun2025interpretabilityparameterspaceminimizing} note that some components partially represent secondary features (See Appendix C2 of \citet{braun2025interpretabilityparameterspaceminimizing}), whereas SPD's masks do not have this issue. Additionally, the neuron contributions of the SPD decomposition are much closer to the target model and have less shrinkage (Compare neuron contributions in Figure 6 of \citet{braun2025interpretabilityparameterspaceminimizing} to our Figure \ref{fig:resid-mlp-weights}). The SPD results also seem to better match theoretical predictions made by a mathematical framework for computation in superposition [unpublished work, forthcoming].

When the importance loss coefficient $\beta_3$ is too small, a single input feature activates a large number of components in $W_\text{in}$ (Figure \ref{fig:resid-mlp-varying-sparsity-importance-vals} - left two columns). This is suboptimal with respect to minimality, and is therefore a poor decomposition. When the importance loss coefficient $\beta_3$ is too high, the importance values decrease below $1$ (Figure \ref{fig:resid-mlp-varying-sparsity-importance-vals} - right two columns), which, aside from being a suboptimal decomposition, is also associated with high reconstruction losses.

\subsection{Toy Models of Cross-Layer Distributed Representations}
\label{sec:residmlp_2layer}
\begin{figure}
    \centering
    \includegraphics[width=0.5\linewidth]{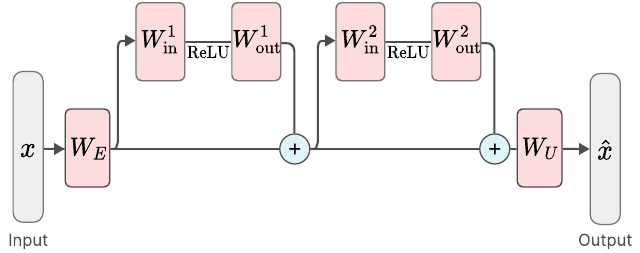}
    \caption{The architecture of one of our two Toy models of Cross-Layer Distributed representations. The other toy model has three MLP blocks instead of two. Figure adapted from \citet{braun2025interpretabilityparameterspaceminimizing}.}
    \label{fig:resid_mlp_2l_architecture}
\end{figure}

Realistic neural networks seem capable of implementing mechanisms distributed across more than one layer \citep{yun2023transformervisualizationdictionarylearning, lindsey2024crosscoders}. To study the ability of SPD to identify components that are spread over multiple layers, \citet{braun2025interpretabilityparameterspaceminimizing} also studied a toy model trained on the same task with the same residual MLP architecture as the one in Section \ref{sec:residmlp_1layer}, but with the $50$ neurons spread over two MLPs instead of one (Figure \ref{fig:resid_mlp_2l_architecture}). 

As in the Toy Model of Compressed Computation, the model learns to compute individual functions using multiple neurons. But here it learns to do so using neurons that are spread over two layers. SPD should find subcomponents in both $W_{\text{in}}^1$ and $W_{\text{in}}^2$ that are causally important for computing each function. And, for the same reasons as in the Toy Model of Compressed Computation,  $W_{\text{out}}^1$ and $W_{\text{out}}^2$ should be decomposed into subcomponents that are in fact all part of one large parameter component, but in this model this one large component should span both layers. 

We apply SPD on this model, as well as another model that spreads $51$ neurons over three MLPs to compute functions of $102$ input features. APD struggled to decompose a model with more than two layers due to hyperparameter sensitivity, but SPD succeeds.

\begin{figure}
    \centering
    \includegraphics[width=0.5\linewidth]{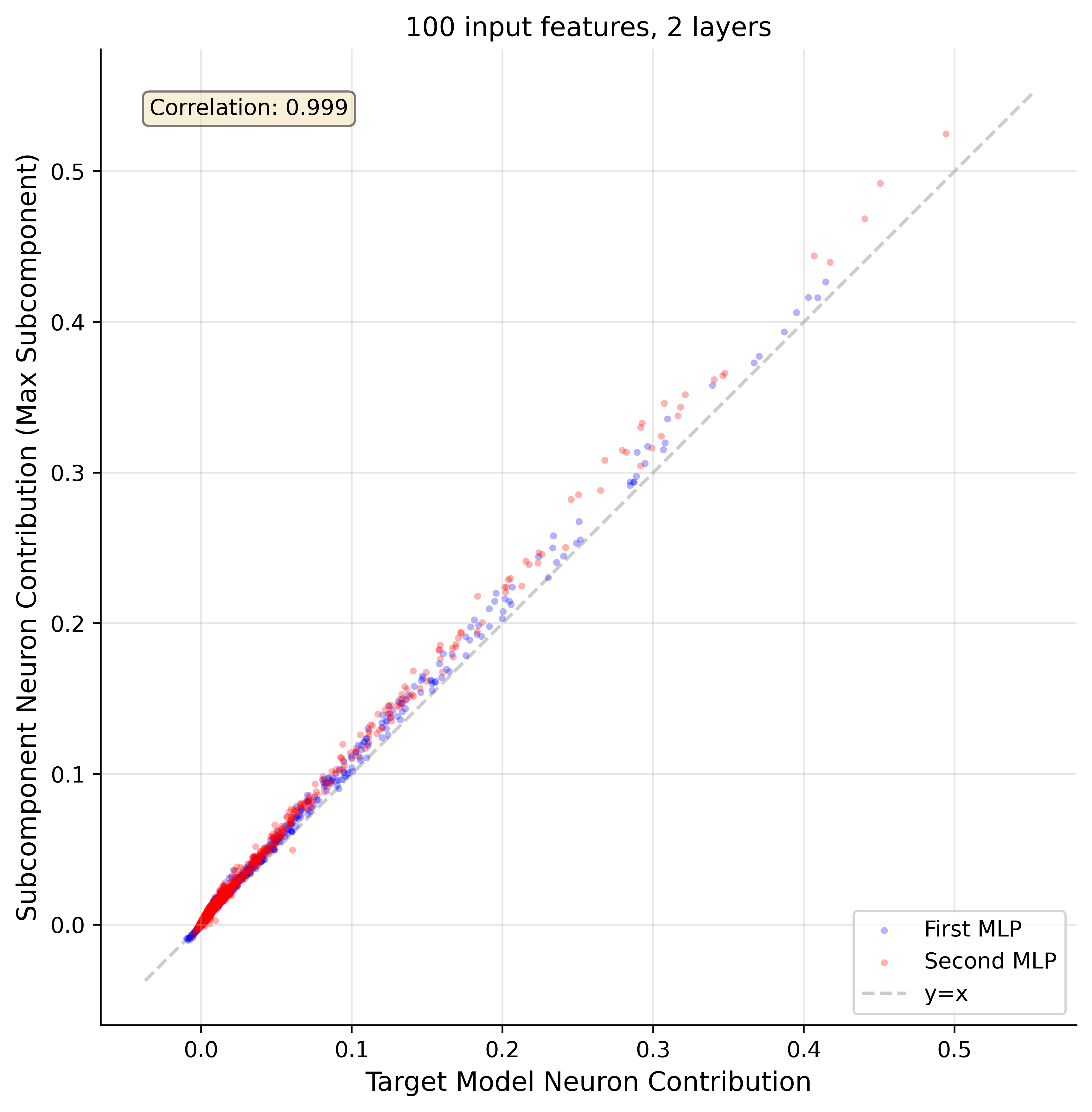}
    \caption{Toy Model of Distributed Representations (Two Layers): Similarity between target model weights and SPD subcomponents for all $100$ input feature dimensions in a $2$-layer residual MLP. Each point represents one neuron's contribution to a particular input feature. \textbf{X-axis}: Neuron contributions measured by Equation \ref{eq:neuron-contribution-model}.  \textbf{Y-axis}: Neuron contributions for the same neuron on the same input feature in the corresponding parameter subcomponents of $W^1_{\text{in}},W^2_{\text{in}}$, measured by Equation \ref{eq:neuron-contribution-subcomponents}. There is a close match between the X and Y axes for each neuron contribution, indicating that each subcomponent connects its corresponding feature to the MLP neurons with similar weights as the target model. However, there is a systematic skew toward higher values on the Y-axis, indicating that the neuron contributions of the subcomponents tend to be slightly larger. This is in contrast to the one-layer case (Figure \ref{fig:resid-mlp-weights-all}) and three-layer case (Figure \ref{fig:resid-mlp-weights-3layers-all}). We currently do not understand the source of this discrepancy, but it is possibly an outcome of suboptimal hyperparameters.}
    \label{fig:resid-mlp-weights-2layers-all}
\end{figure}

\subsubsection*{SPD Results for two- and three-layer models of Cross-Layer Distributed Representations}
SPD finds qualitatively similar results to the $1$-layer Toy Model of Compressed Computation presented in Section \ref{sec:residmlp_1layer}. In the two-layer model, the MLP input matrices $W^1_\text{in}, W^2_\text{in}$, which each have shape ($\text{d\_mlp}\mathbin{=}25$, $\text{d\_embed}\mathbin{=}1000$), are decomposed into $100$ components that each compute the functions for one of the input features, using neurons spread over both MLP layers. The MLP output matrices $W^1_\text{out}, W^2_\text{out}$ are decomposed into a single rank $50$ component. Notably, however, this component is now spread over both layers (Figure \ref{fig:resid-mlp-importances-2layers}). This demonstrates that SPD can identify mechanisms that span multiple layers.

\begin{figure}
    \centering
    \includegraphics[width=0.5\linewidth]{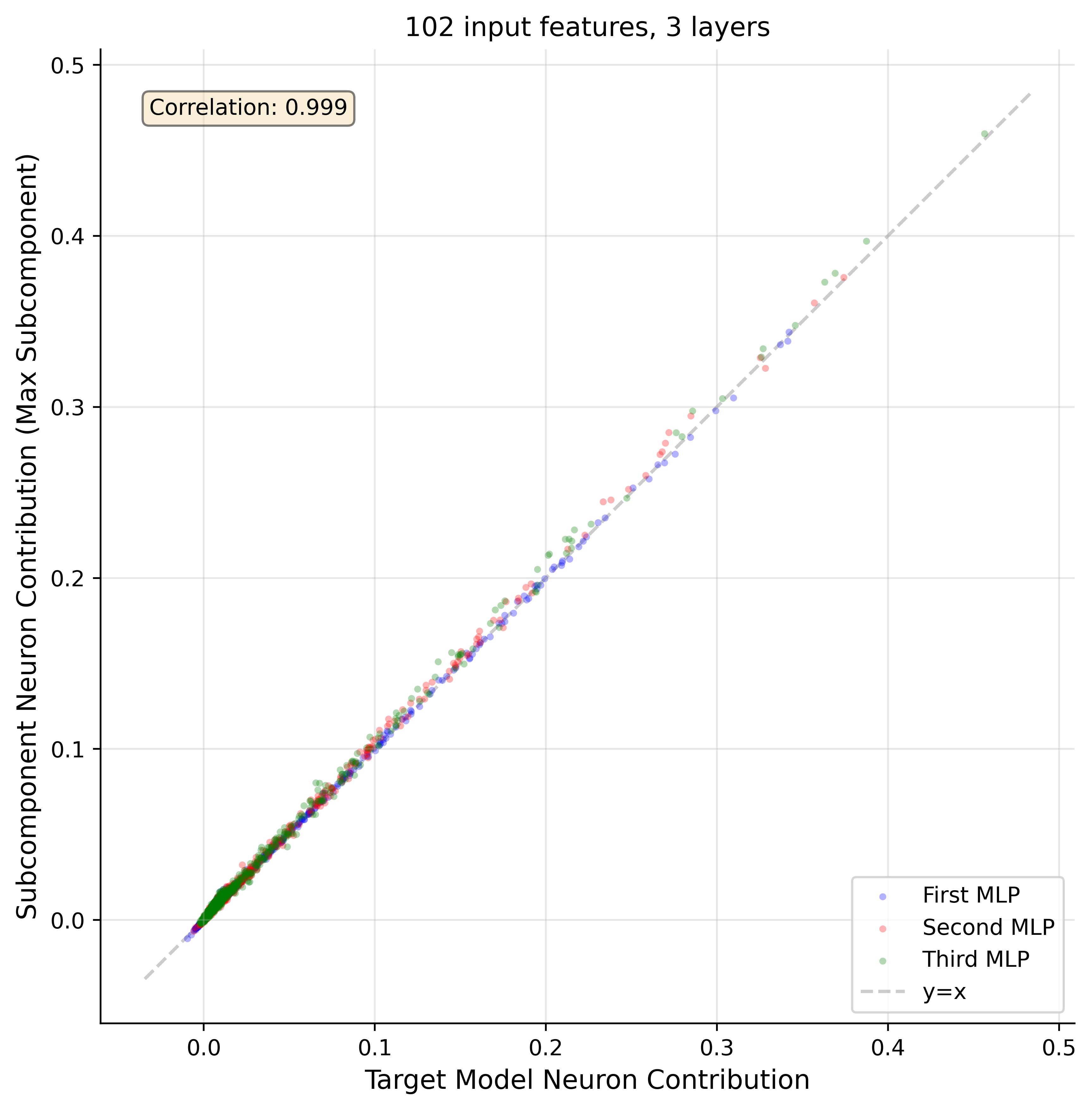}
    \caption{Toy Model of Distributed Representations (Three Layers): Similarity between target model weights and SPD subcomponents for all $102$ input feature dimensions in a $3$-layer residual MLP. Each point represents one neuron's contribution for a particular input feature. \textbf{X-axis}: Neuron contributions measured by Equation \ref{eq:neuron-contribution-model}.  \textbf{Y-axis}: Neuron contributions for the same neuron on the same input feature in the corresponding parameter subcomponents of $W^1_{\text{in}},W^2_{\text{in}}, W^3_{\text{in}}$, measured by Equation \ref{eq:neuron-contribution-subcomponents}.  There is a close match between the X and Y axes for each neuron contribution, indicating that each subcomponent connects its corresponding feature to the MLP neurons with similar weights as the target model.}
    \label{fig:resid-mlp-weights-3layers-all}
\end{figure}

\begin{figure}
    \centering
    \includegraphics[width=0.35\linewidth]{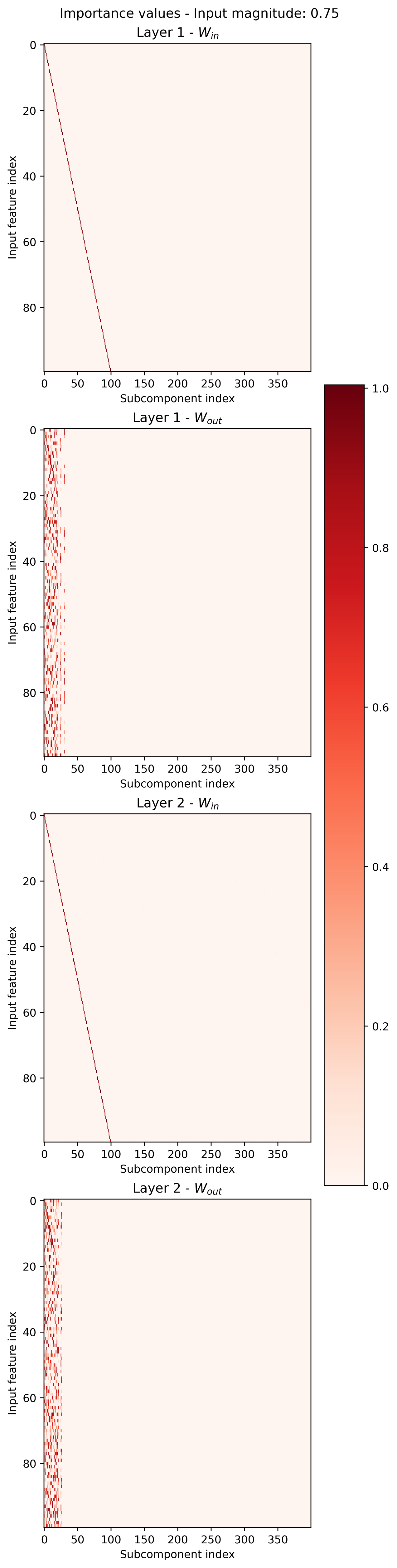}
    \caption{Toy Model of Distributed Representations (Two Layers): Causal importance values of each subcomponent (clipped between $0$ and $1$) for the matrices in both MLP layers, $W^1_{\text{in}},W^1_{\text{out}},W^2_{\text{in}},W^2_{\text{out}}$ in response to one-hot inputs ($x_i=0.75$). Each subcomponent in $W^1_{\text{in}},W^2_{\text{in}}$ has causal importance for computing a unique input feature. On the other hand, the combined $50$ subcomponents of $W^1_{\text{out}},W^2_{\text{out}}$ all coactivate for all input features, indicating they are part of a single rank-$50$ identity component. The columns of each matrix are permuted differently, ordered by iteratively choosing the subcomponent (without replacement) with the highest causal importance for each input feature. }
    \label{fig:resid-mlp-importances-2layers}
\end{figure}

\begin{figure}
    \centering
    \includegraphics[width=0.23\linewidth]{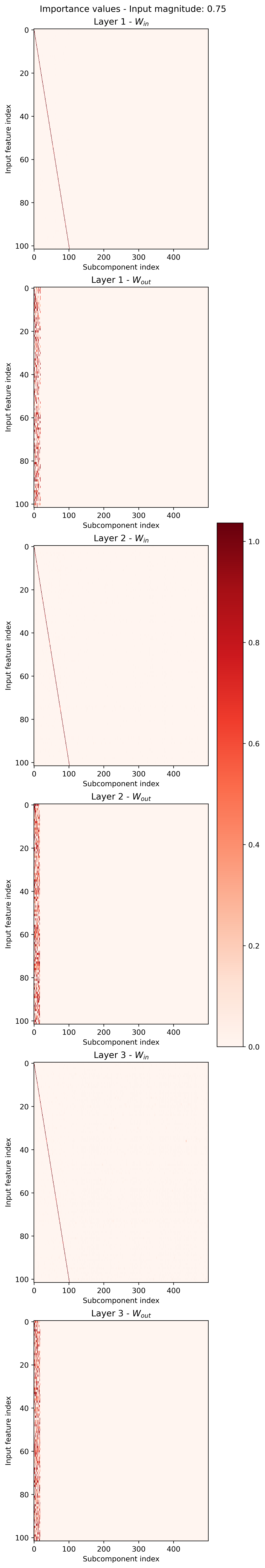}
    \caption{Toy Model of Distributed Representations (Three Layers): Causal importance values of each subcomponent (clipped between $0$ and $1$) for the matrices in all three MLP layers, $W^1_{\text{in}},W^1_{\text{out}},W^2_{\text{in}},W^2_{\text{out}},W^3_{\text{in}},W^3_{\text{out}}$ in response to one-hot inputs ($x_i=0.75$). Each subcomponent in $W^1_{\text{in}},W^2_{\text{in}},W^3_{\text{in}}$ has causal importance for computing a unique input feature. On the other hand, the combined $51$ subcomponents of $W^1_{\text{out}},W^2_{\text{out}},W^3_{\text{out}}$ all coactivate for all input features, indicating they are part of a single rank-$51$ identity component. The columns of each matrix are permuted differently, ordered by iteratively choosing the subcomponent (without replacement) with the highest causal importance for each input feature.}
    \label{fig:resid-mlp-importances-3layers}
\end{figure}

In the three-layer model, the MLP input matrices $W^1_\text{in}, W^2_\text{in}, W^3_\text{in}$, which each have shape ($\text{d\_mlp}\mathbin{=}17$, $\text{d\_embed}\mathbin{=}1000$), are likewise decomposed into $102$ components all spread over 3 layers, and the MLP output matrices $W^1_\text{out}, W^2_\text{out}, W^3_\text{out}$ are decomposed into a single rank $51$ component spread over all three layers (Figure \ref{fig:resid-mlp-importances-3layers}).
In both the two- and three-layer models, we find that the computations occurring in each parameter subcomponent of the $W_{in}$ matrices closely correspond to individual input feature computations in the target model (Figures  \ref{fig:resid-mlp-weights-2layers-all}, \ref{fig:resid-mlp-weights-3layers-all}, \ref{fig:resid-mlp-weights-2layers}, \ref{fig:resid-mlp-weights-3layers}). We note, however, that there is sensitivity to random seeds in these results for these models, possibly due to unfavorable random initialization, which we intend to explore in future work. 


\section{Related Work}\label{sec:related-work}

SPD uses randomly sampled masks in order to identify the minimal set of parameter subcomponents that are causally important for computing a model's output on a given input. This is, in essence, a method for finding a good set of causal mediators \citep{mueller2024questrightmediatorhistory, vig2020causalmediationanalysisinterpreting, geiger2021causalabstractionsneuralnetworks, geiger2024findingalignmentsinterpretablecausal}: Our causal importance function can be thought of as learning how to causally intervene on our target model in order to identify a minimal set of simple causal mediators of a model's computation. However, unlike many approaches in the causal intervention literature (e.g. \citet{chan2022causalscrubbing, wang2022interpretability, conmy2023automatedcircuitdiscoverymechanistic}), our approach does not assume a particular basis for these interventions. Instead, it learns the basis in which causal interventions are made. It also makes these interventions in parameter space. 

Some previous work learns fixed masks to ablate parts of the model's input or parts of its parameters \citep{csordás2021neuralnetsmodularinspecting, cao2021lowcomplexityprobingfindingsubnetworks, zhang2021subnetworkstructurekeyoutofdistribution}. But these masks tend to be for fixed inputs, and also often assume a particular basis. Our work also has an important difference to other masking-based attribution approaches: While our causal importances are predicted, the masks themselves are stochastically sampled with an amount of randomness based on the predicted causal importances.

Our definition of causal importance of subcomponents is related but not identical to the definition of causal dependence used in other literature \citep{Lewis1973-LEWC-2, mueller2024missedcausesambiguouseffects}. For one, it may be possible that networks compute particular outputs even if one causally important subcomponent is ablated, using emergent self-repair \citep{mcgrath2023hydraeffectemergentselfrepair}. In that case, a subcomponent may be causally important for a particular output, but the output may not be causally dependent on it.

SPD also has parallels to work that uses attribution methods to approximate the causal importance of model components (e.g. \citet{mozer1988skeletonization, syed2023attributionpatchingoutperformsautomated, marks2024sparsefeaturecircuitsdiscovering, braun2025interpretabilityparameterspaceminimizing}) or the inputs (as in saliency maps) (e.g. \citet{Fong_2017_MaskingBasedCausalAttribution, simonyan2014deepinsideconvolutionalnetworks}). Our approach learns to predict causal importances given an input, which can be thought of as learning to predict attributions. To the best of our knowledge, we are not aware of similar approaches that learn to predict the input-dependent ablatability of model components using the method described in this paper.

\citet{chrisman2025identifyingsparselyactivecircuits} decomposed networks in parameter space by finding low-rank parameter components that can reconstruct, using sparse coefficients, the gradient of a loss between the network output and a baseline output. Somewhat similarly, \citet{matena2025uncoveringmodelprocessingstrategies} decomposed models in parameter space via non-negative factorisation of the models' per-sample Fisher Information matrices into components. SPD also decomposes networks into low-rank components in parameter space based on how the network output responds to perturbations. But instead of relying on local approximations like gradients to estimate the effects of a perturbation, it is trained by directly checking the effect ablating components in various combinations has on the output.

SPD can be viewed as approximately quantifying the degeneracy in neural network weights over different subdistributions of the data. SPD was in part inspired by singular learning theory \citep{watanabe2009algebraic}, which quantifies degeneracy in network weights present over the entire data distribution using the learning coefficient. \citet{wang2024differentiationspecializationattentionheads} defined the data-refined learning coefficient, which measures degeneracy in neural network weights over a chosen subset of the distribution. 
In contrast, SPD works in an unsupervised manner, finding a single set of vectors to represent the neural network parameters, such that as many vectors as possible are degenerate on any given datapoint in the distribution. SPD also requires vectors to be ablatable to zero rather than just being degenerate locally.

See \citet{braun2025interpretabilityparameterspaceminimizing} for further discussion on the relationship between linear parameter decomposition methods and sparse autoencoders; transcoders; weight masking and pruning; circuit discovery and causal mediation analysis; interpretability of neural network parameters; mixture of experts; and loss landscape intrinsic dimensionality and degeneracy.

\section{Discussion}\label{sec:discussion}

In this paper we introduced SPD, a method that resolves many of the issues of APD \citep{braun2025interpretabilityparameterspaceminimizing}. The method is considerably more scalable and robust to hyperparameters, which we demonstrate by using the method to decompose deeper and more complex models than APD has successfully decomposed. 

We hypothesize that this relative robustness comes from various sources: 
\begin{enumerate}
    \item \textbf{APD required estimating the expected number of active components in advance}, because it needed to set the hyperparameter $k$ for selecting the top-$k$ most attributed components per batch. This number would usually not be known in advance for realistic models. APD results were very sensitive to it. SPD uses trained causal importance functions instead, and therefore no longer needs to use a fixed estimate for the number of active subcomponents per datapoint. We still need to pick the loss coefficient for the causal importance penalty $\beta_3$, but this is a much more forgiving hyperparameter than the hyper-sensitive top-$k$ hyperparameter\footnote{In an idealized setting where (a) there is no batch noise in the loss, (b) no finite floating point precision, and (c) inactive mechanisms do not causally influence the network output at all, then setting $\beta_3$ to any value infinitesimally larger than zero should suffice to make our desired decomposition the global optimum.}
    \item \textbf{Gradients flow through every subcomponent on every datapoint}, unlike in APD, where gradients only flowed through the top-$k$ most attributed components. Top-$k$ activation functions create discontinuities that, in general, tend to lead to unstable gradient-based training. In SPD, even subcomponents with causal importance values of zero will almost always permit gradients to flow, thus helping them error-correct if they are wrong. 
    \item \textbf{SPD does not need to optimize for `simplicity'} (in the sense of \citet{braun2025interpretabilityparameterspaceminimizing}, where simple parameter components span as few ranks and layers as possible). Not only does this remove one hyperparameter (making tuning easier), but it also avoids inducing shrinkage in the singular values of the parameter component weight matrices. SPD does not exhibit shrinkage in the parameter subcomponents because the importance norm penalizes the probability that a subcomponent will be unmasked, but does not directly penalize norms of the singular values of the parameter matrices themselves. This can be helpful for learning correct solutions: For example, if correctly-oriented parameter components exhibit shrinkage, then their sum will not sum to the parameters of the target model, and therefore other parameter components will need to compensate. For this reason, the faithfulness and simplicity losses in APD were in tension. Removing this tension, and removing a whole hyperparameter to tune, makes it easier to hit small targets in parameter space. Although there is shrinkage in the causal importance values, this does not appear to be very influential since causal importance values only determine the allowed minimum value of the masks. For example, a causal importance value of $0.95$ indicates that we can ablate a subcomponent by up to five percent without significantly affecting the network output, but we can also just not ablate it at all. %
    \item \textbf{APD used gradient-based attribution methods to estimate causal importance}, even though those methods are only first-order approximations of ideal causal attributions, which is often a poor approximation \citep{watson2022conceptual, kramár2024atpefficientscalablemethod}. If causal attributions are wrong, then the wrong parameter components would activate, so the wrong parameter components would be trained to compute the model's function on particular inputs. Systematic errors in causal attributions will lead to systematic biases in the gradients, which can be catastrophic when trying to hit a very particular target in parameter space. SPD instead directly optimizes for causal importance and is likely a much better estimate than the approximations found by even somewhat sophisticated attribution methods (e.g. \citet{sundararajan2017axiomatic}).

\end{enumerate}

We hope that the stability and scalability of SPD will facilitate further scaling to much larger models than the ones studied here. In future work, we plan to test the scaling limits of the current method and explore any necessary adjustments for increased scalability and robustness.  

We plan to investigate several outlying issues in future work. One issue is that we are unsure if learning independent subcomponents will enable the method to learn subcomponents that can describe the network's function using as short a description as possible. It may be possible that information from future layers is necessary to identify whether a given subcomponent is causally important. If it is, then calculating causal importance values layerwise will mean that some subcomponents are active when they need not be. It may therefore be interesting to explore causal importance functions that take as input more global information from throughout the network, rather than only the subcomponent inner activations at a given layer. Another issue is that both APD and SPD privilege mechanisms that span individual layers due to the importance loss or simplicity loss in SPD and APD respectively; it may be desirable to identify loss functions that privilege layers less.  

The toy models in our work had known ground truth mechanisms, and therefore it was straightforward to identify which subcomponents should be grouped together into full parameter components. However, in the general case we will not know this by default. We therefore need to develop approaches that cluster subcomponents together in a way that combines the sparse and dense coding schemes laid out in the appendix of \citet{braun2025interpretabilityparameterspaceminimizing} to achieve components that permit a minimum length description of the network's function in terms of parameter components. 

It is worth noting that the SPD approach can be generalized in multiple straightforward ways. It is not necessary, for instance, to decompose parameter components strictly into subcomponents consisting of rank-one matrices. Subcomponents could, for instance, span only one rank but across all matrices in all layers. Alternatively, different implementations of the causal importance function could be used.

We expect that SPD's scalability and stability will enable new research directions previously inaccessible with APD, such as investigating mechanisms of memorization and their relationship to neural network parameter storage capacity. Additionally, the principles behind SPD may be useful for training intrinsically decomposed models.

\section*{Acknowledgments}

We thank Tom McGrath, Stefan Heimersheim, Daniel Filan, Bart Bussmann, Logan Smith, Nathan Hu, Dashiell Stander, Brianna Chrisman, Kola Ayonrinde, and Atticus Geiger for helpful feedback on previous drafts of this paper. We also thank Stefan Heimersheim for initially suggesting the idea to train on an early version of what became the stochastic loss, which we had up to then only treated as a validation metric. Additionally, we thank Kaarel Hänni, whose inputs helped us develop the definition of component causal importance, and Linda Linsefors, whose corrections of earlier work on interference terms arising in computation in superposition helped us interpret the toy model of compressed computation.

\begin{raggedright}
\bibliography{references}

\begin{thebibliography}{45}
\providecommand{\natexlab}[1]{#1}
\providecommand{\url}[1]{\texttt{#1}}
\expandafter\ifx\csname urlstyle\endcsname\relax
  \providecommand{\doi}[1]{doi: #1}\else
  \providecommand{\doi}{doi: \begingroup \urlstyle{rm}\Url}\fi

\bibitem[Adebayo et~al.(2018)Adebayo, Gilmer, Muelly, Goodfellow, Hardt, and Kim]{adebayo2020sanitycheckssaliencymaps}
Julius Adebayo, Justin Gilmer, Michael Muelly, Ian Goodfellow, Moritz Hardt, and Been Kim.
\newblock Sanity checks for saliency maps.
\newblock In S.~Bengio, H.~Wallach, H.~Larochelle, K.~Grauman, N.~Cesa-Bianchi, and R.~Garnett, editors, \emph{Advances in Neural Information Processing Systems}, volume~31. Curran Associates, Inc., 2018.
\newblock URL \url{https://proceedings.neurips.cc/paper_files/paper/2018/file/294a8ed24b1ad22ec2e7efea049b8737-Paper.pdf}.

\bibitem[Bhagat et~al.(2025)Bhagat, Medina, Giglemiani, and Heimersheim]{BhagatCC}
Jai Bhagat, Sara~Molas Medina, Giorgi Giglemiani, and Stefan Heimersheim.
\newblock Compressed computation is (probably) not computation in superposition.
\newblock \url{https://www.lesswrong.com/posts/ZxFchCFJFcgysYsT9/compressed-computation-is-probably-not-computation-in}, 2025.

\bibitem[Braun et~al.(2025)Braun, Bushnaq, Heimersheim, Mendel, and Sharkey]{braun2025interpretabilityparameterspaceminimizing}
Dan Braun, Lucius Bushnaq, Stefan Heimersheim, Jake Mendel, and Lee Sharkey.
\newblock Interpretability in parameter space: Minimizing mechanistic description length with attribution-based parameter decomposition, 2025.
\newblock URL \url{https://arxiv.org/abs/2501.14926}.

\bibitem[Bricken et~al.(2023)Bricken, Templeton, Batson, Chen, Jermyn, Conerly, Turner, Anil, Denison, Askell, Lasenby, Wu, Kravec, Schiefer, Maxwell, Joseph, Hatfield-Dodds, Tamkin, Nguyen, McLean, Burke, Hume, Carter, Henighan, and Olah]{bricken2023monosemanticity}
Trenton Bricken, Adly Templeton, Joshua Batson, Brian Chen, Adam Jermyn, Tom Conerly, Nick Turner, Cem Anil, Carson Denison, Amanda Askell, Robert Lasenby, Yifan Wu, Shauna Kravec, Nicholas Schiefer, Tim Maxwell, Nicholas Joseph, Zac Hatfield-Dodds, Alex Tamkin, Karina Nguyen, Brayden McLean, Josiah~E Burke, Tristan Hume, Shan Carter, Tom Henighan, and Christopher Olah.
\newblock Towards monosemanticity: Decomposing language models with dictionary learning.
\newblock \emph{Transformer Circuits Thread}, 2023.
\newblock URL \url{https://transformer-circuits.pub/2023/monosemantic-features/index.html}.

\bibitem[Cao et~al.(2021)Cao, Sanh, and Rush]{cao2021lowcomplexityprobingfindingsubnetworks}
Steven Cao, Victor Sanh, and Alexander~M. Rush.
\newblock Low-complexity probing via finding subnetworks, 2021.
\newblock URL \url{https://arxiv.org/abs/2104.03514}.

\bibitem[Chan et~al.(2022)Chan, Garriga-Alonso, Goldowsky-Dill, Greenblatt, Nitishinskaya, Radhakrishnan, Shlegeris, and Thomas]{chan2022causalscrubbing}
Lawrence Chan, Adrià Garriga-Alonso, Nicholas Goldowsky-Dill, Ryan Greenblatt, Jenny Nitishinskaya, Ansh Radhakrishnan, Buck Shlegeris, and Nate Thomas.
\newblock Causal scrubbing: a method for rigorously testing interpretability hypotheses [redwood research], December 2022.
\newblock URL \url{https://www.alignmentforum.org/posts/JvZhhzycHu2Yd57RN/causal-scrubbing-a-method-for-rigorously-testing}.

\bibitem[Chanin et~al.(2024)Chanin, Wilken-Smith, Dulka, Bhatnagar, and Bloom]{chanin2024absorptionstudyingfeaturesplitting}
David Chanin, James Wilken-Smith, Tomáš Dulka, Hardik Bhatnagar, and Joseph Bloom.
\newblock A is for absorption: Studying feature splitting and absorption in sparse autoencoders, 2024.
\newblock URL \url{https://arxiv.org/abs/2409.14507}.

\bibitem[Chrisman et~al.(2025)Chrisman, Bushnaq, and Sharkey]{chrisman2025identifyingsparselyactivecircuits}
Brianna Chrisman, Lucius Bushnaq, and Lee Sharkey.
\newblock Identifying sparsely active circuits through local loss landscape decomposition, 2025.
\newblock URL \url{https://arxiv.org/abs/2504.00194}.

\bibitem[Conmy et~al.(2023)Conmy, Mavor-Parker, Lynch, Heimersheim, and Garriga-Alonso]{conmy2023automatedcircuitdiscoverymechanistic}
Arthur Conmy, Augustine Mavor-Parker, Aengus Lynch, Stefan Heimersheim, and Adri\`{a} Garriga-Alonso.
\newblock Towards automated circuit discovery for mechanistic interpretability.
\newblock In A.~Oh, T.~Naumann, A.~Globerson, K.~Saenko, M.~Hardt, and S.~Levine, editors, \emph{Advances in Neural Information Processing Systems}, volume~36, pages 16318--16352. Curran Associates, Inc., 2023.
\newblock URL \url{https://proceedings.neurips.cc/paper_files/paper/2023/file/34e1dbe95d34d7ebaf99b9bcaeb5b2be-Paper-Conference.pdf}.

\bibitem[Csord{\'a}s et~al.(2021)Csord{\'a}s, van Steenkiste, and Schmidhuber]{csordás2021neuralnetsmodularinspecting}
R{\'o}bert Csord{\'a}s, Sjoerd van Steenkiste, and J{\"u}rgen Schmidhuber.
\newblock Are neural nets modular? inspecting functional modularity through differentiable weight masks.
\newblock In \emph{International Conference on Learning Representations}, 2021.
\newblock URL \url{https://openreview.net/forum?id=7uVcpu-gMD}.

\bibitem[Cunningham et~al.(2024)Cunningham, Smith, Ewart, Huben, and Sharkey]{cunningham2023sparseautoencodershighlyinterpretable}
Hoagy Cunningham, Logan~Riggs Smith, Aidan Ewart, Robert Huben, and Lee Sharkey.
\newblock Sparse autoencoders find highly interpretable features in language models.
\newblock In \emph{The Twelfth International Conference on Learning Representations}, 2024.
\newblock URL \url{https://openreview.net/forum?id=F76bwRSLeK}.

\bibitem[Elhage et~al.(2022)Elhage, Hume, Olsson, Schiefer, Henighan, Kravec, Hatfield-Dodds, Lasenby, Drain, Chen, Grosse, McCandlish, Kaplan, Amodei, Wattenberg, and Olah]{elhage2022toy}
Nelson Elhage, Tristan Hume, Catherine Olsson, Nicholas Schiefer, Tom Henighan, Shauna Kravec, Zac Hatfield-Dodds, Robert Lasenby, Dawn Drain, Carol Chen, Roger Grosse, Sam McCandlish, Jared Kaplan, Dario Amodei, Martin Wattenberg, and Christopher Olah.
\newblock Toy models of superposition, 2022.

\bibitem[Fong and Vedaldi(2017)]{Fong_2017_MaskingBasedCausalAttribution}
Ruth~C. Fong and Andrea Vedaldi.
\newblock Interpretable explanations of black boxes by meaningful perturbation.
\newblock In \emph{2017 IEEE International Conference on Computer Vision (ICCV)}. IEEE, October 2017.
\newblock \doi{10.1109/iccv.2017.371}.
\newblock URL \url{http://dx.doi.org/10.1109/ICCV.2017.371}.

\bibitem[Gao et~al.(2024)Gao, la~Tour, Tillman, Goh, Troll, Radford, Sutskever, Leike, and Wu]{gao2024scalingevaluatingsparseautoencoders}
Leo Gao, Tom~Dupré la~Tour, Henk Tillman, Gabriel Goh, Rajan Troll, Alec Radford, Ilya Sutskever, Jan Leike, and Jeffrey Wu.
\newblock Scaling and evaluating sparse autoencoders, 2024.
\newblock URL \url{https://arxiv.org/abs/2406.04093}.

\bibitem[Geiger et~al.(2024{\natexlab{a}})Geiger, Lu, Icard, and Potts]{geiger2021causalabstractionsneuralnetworks}
Atticus Geiger, Hanson Lu, Thomas Icard, and Christopher Potts.
\newblock Causal abstractions of neural networks.
\newblock In \emph{Proceedings of the 35th International Conference on Neural Information Processing Systems}, NIPS '21, Red Hook, NY, USA, 2024{\natexlab{a}}. Curran Associates Inc.
\newblock ISBN 9781713845393.

\bibitem[Geiger et~al.(2024{\natexlab{b}})Geiger, Wu, Potts, Icard, and Goodman]{geiger2024findingalignmentsinterpretablecausal}
Atticus Geiger, Zhengxuan Wu, Christopher Potts, Thomas Icard, and Noah Goodman.
\newblock Finding alignments between interpretable causal variables and distributed neural representations.
\newblock In Francesco Locatello and Vanessa Didelez, editors, \emph{Proceedings of the Third Conference on Causal Learning and Reasoning}, volume 236 of \emph{Proceedings of Machine Learning Research}, pages 160--187. PMLR, 01--03 Apr 2024{\natexlab{b}}.
\newblock URL \url{https://proceedings.mlr.press/v236/geiger24a.html}.

\bibitem[Hänni et~al.(2024)Hänni, Mendel, Vaintrob, and Chan]{hänni2024mathematicalmodelscomputationsuperposition}
Kaarel Hänni, Jake Mendel, Dmitry Vaintrob, and Lawrence Chan.
\newblock Mathematical models of computation in superposition, 2024.
\newblock URL \url{https://arxiv.org/abs/2408.05451}.

\bibitem[Jermyn et~al.(2024)Jermyn, Templeton, Batson, and Bricken]{jermyn2024tanh}
Adam Jermyn, Adly Templeton, Joshua Batson, and Trenton Bricken.
\newblock Tanh penalty in dictionary learning.
\newblock \url{https://transformer-circuits.pub/2024/feb-update/index.html#:~:text=handle%20dying%20neurons.-,Tanh%20Penalty%20in%20Dictionary%20Learning,-Adam%20Jermyn%2C%20Adly}, 2024.

\bibitem[Kingma and Welling(2013)]{kingma2013autoencodingvariationalbayes}
Diederik~P Kingma and Max Welling.
\newblock Auto-encoding variational bayes, 2013.
\newblock URL \url{https://arxiv.org/abs/1312.6114}.

\bibitem[Kramár et~al.(2024)Kramár, Lieberum, Shah, and Nanda]{kramár2024atpefficientscalablemethod}
János Kramár, Tom Lieberum, Rohin Shah, and Neel Nanda.
\newblock Atp*: An efficient and scalable method for localizing llm behaviour to components, 2024.
\newblock URL \url{https://arxiv.org/abs/2403.00745}.

\bibitem[Leask et~al.(2025)Leask, Bussmann, Pearce, Bloom, Tigges, Moubayed, Sharkey, and Nanda]{leask2025sparseautoencoderscanonicalunits}
Patrick Leask, Bart Bussmann, Michael Pearce, Joseph Bloom, Curt Tigges, Noura~Al Moubayed, Lee Sharkey, and Neel Nanda.
\newblock Sparse autoencoders do not find canonical units of analysis, 2025.
\newblock URL \url{https://arxiv.org/abs/2502.04878}.

\bibitem[Lewis(1973)]{Lewis1973-LEWC-2}
David~K. Lewis.
\newblock \emph{Counterfactuals}.
\newblock Blackwell, Malden, Mass., 1973.

\bibitem[Lindsay et~al.(2024)Lindsay, Templeton, Marcus, Conerly, Batson, and Olah]{lindsey2024crosscoders}
Jack Lindsay, Adly Templeton, Jonathan Marcus, Thomas Conerly, Joshua Batson, and Christopher Olah.
\newblock Sparse crosscoders for cross-layer features and model diffing, October 2024.
\newblock URL \url{https://transformer-circuits.pub/2024/crosscoders/index.html}.

\bibitem[Loshchilov and Hutter(2019)]{loshchilov2019adamw}
Ilya Loshchilov and Frank Hutter.
\newblock Decoupled weight decay regularization, 2019.
\newblock URL \url{https://arxiv.org/abs/1711.05101}.

\bibitem[Marks et~al.(2024)Marks, Rager, Michaud, Belinkov, Bau, and Mueller]{marks2024sparsefeaturecircuitsdiscovering}
Samuel Marks, Can Rager, Eric~J. Michaud, Yonatan Belinkov, David Bau, and Aaron Mueller.
\newblock Sparse feature circuits: Discovering and editing interpretable causal graphs in language models, 2024.
\newblock URL \url{https://arxiv.org/abs/2403.19647}.

\bibitem[Matena and Raffel(2025)]{matena2025uncoveringmodelprocessingstrategies}
Michael Matena and Colin Raffel.
\newblock Uncovering model processing strategies with non-negative per-example fisher factorization, 2025.
\newblock URL \url{https://arxiv.org/abs/2310.04649}.

\bibitem[McGrath et~al.(2023)McGrath, Rahtz, Kramar, Mikulik, and Legg]{mcgrath2023hydraeffectemergentselfrepair}
Thomas McGrath, Matthew Rahtz, Janos Kramar, Vladimir Mikulik, and Shane Legg.
\newblock The hydra effect: Emergent self-repair in language model computations, 2023.
\newblock URL \url{https://arxiv.org/abs/2307.15771}.

\bibitem[Mendel(2024)]{mendel2024sae}
Jake Mendel.
\newblock {SAE} feature geometry is outside the superposition hypothesis.
\newblock \emph{Alignment Forum}, 2024.
\newblock URL \url{https://www.alignmentforum.org/posts/MFBTjb2qf3ziWmzz6/sae-feature-geometry-is-outside-the-superposition-hypothesis}.

\bibitem[Mozer and Smolensky(1988)]{mozer1988skeletonization}
Michael~C Mozer and Paul Smolensky.
\newblock Skeletonization: A technique for trimming the fat from a network via relevance assessment.
\newblock In D.~Touretzky, editor, \emph{Advances in Neural Information Processing Systems}, volume~1. Morgan-Kaufmann, 1988.
\newblock URL \url{https://proceedings.neurips.cc/paper_files/paper/1988/file/07e1cd7dca89a1678042477183b7ac3f-Paper.pdf}.

\bibitem[Mueller(2024)]{mueller2024missedcausesambiguouseffects}
Aaron Mueller.
\newblock Missed causes and ambiguous effects: Counterfactuals pose challenges for interpreting neural networks, 2024.
\newblock URL \url{https://arxiv.org/abs/2407.04690}.

\bibitem[Mueller et~al.(2024)Mueller, Brinkmann, Li, Marks, Pal, Prakash, Rager, Sankaranarayanan, Sharma, Sun, Todd, Bau, and Belinkov]{mueller2024questrightmediatorhistory}
Aaron Mueller, Jannik Brinkmann, Millicent Li, Samuel Marks, Koyena Pal, Nikhil Prakash, Can Rager, Aruna Sankaranarayanan, Arnab~Sen Sharma, Jiuding Sun, Eric Todd, David Bau, and Yonatan Belinkov.
\newblock The quest for the right mediator: A history, survey, and theoretical grounding of causal interpretability, 2024.
\newblock URL \url{https://arxiv.org/abs/2408.01416}.

\bibitem[Sharkey et~al.(2022)Sharkey, Braun, and Millidge]{Sharkey_Braun_Millidge_2022}
Lee Sharkey, Dan Braun, and Beren Millidge.
\newblock Taking features out of superposition with sparse autoencoders, Dec 2022.
\newblock URL \url{https://www.alignmentforum.org/posts/z6QQJbtpkEAX3Aojj/interim-research-report-taking-features-out-of-superposition}.

\bibitem[Sharkey et~al.(2025)Sharkey, Chughtai, Batson, Lindsey, Wu, Bushnaq, Goldowsky-Dill, Heimersheim, Ortega, Bloom, Biderman, Garriga-Alonso, Conmy, Nanda, Rumbelow, Wattenberg, Schoots, Miller, Michaud, Casper, Tegmark, Saunders, Bau, Todd, Geiger, Geva, Hoogland, Murfet, and McGrath]{sharkey2025openproblemsmechanisticinterpretability}
Lee Sharkey, Bilal Chughtai, Joshua Batson, Jack Lindsey, Jeff Wu, Lucius Bushnaq, Nicholas Goldowsky-Dill, Stefan Heimersheim, Alejandro Ortega, Joseph Bloom, Stella Biderman, Adria Garriga-Alonso, Arthur Conmy, Neel Nanda, Jessica Rumbelow, Martin Wattenberg, Nandi Schoots, Joseph Miller, Eric~J. Michaud, Stephen Casper, Max Tegmark, William Saunders, David Bau, Eric Todd, Atticus Geiger, Mor Geva, Jesse Hoogland, Daniel Murfet, and Tom McGrath.
\newblock Open problems in mechanistic interpretability, 2025.
\newblock URL \url{https://arxiv.org/abs/2501.16496}.

\bibitem[Simonyan et~al.(2014)Simonyan, Vedaldi, and Zisserman]{simonyan2014deepinsideconvolutionalnetworks}
Karen Simonyan, Andrea Vedaldi, and Andrew Zisserman.
\newblock Deep inside convolutional networks: Visualising image classification models and saliency maps, 2014.
\newblock URL \url{https://arxiv.org/abs/1312.6034}.

\bibitem[Sundararajan et~al.(2017)Sundararajan, Taly, and Yan]{sundararajan2017axiomatic}
Mukund Sundararajan, Ankur Taly, and Qiqi Yan.
\newblock Axiomatic attribution for deep networks, 2017.

\bibitem[Syed et~al.(2024)Syed, Rager, and Conmy]{syed2023attributionpatchingoutperformsautomated}
Aaquib Syed, Can Rager, and Arthur Conmy.
\newblock Attribution patching outperforms automated circuit discovery.
\newblock In Yonatan Belinkov, Najoung Kim, Jaap Jumelet, Hosein Mohebbi, Aaron Mueller, and Hanjie Chen, editors, \emph{Proceedings of the 7th BlackboxNLP Workshop: Analyzing and Interpreting Neural Networks for NLP}, pages 407--416, Miami, Florida, US, November 2024. Association for Computational Linguistics.
\newblock \doi{10.18653/v1/2024.blackboxnlp-1.25}.
\newblock URL \url{https://aclanthology.org/2024.blackboxnlp-1.25/}.

\bibitem[Till(2024)]{till2024truefeatures}
Demian Till.
\newblock Do sparse autoencoders find "true features"?, February 2024.
\newblock URL \url{https://www.lesswrong.com/posts/QoR8noAB3Mp2KBA4B/do-sparse-autoencoders-find-true-features}.

\bibitem[Vig et~al.(2020)Vig, Gehrmann, Belinkov, Qian, Nevo, Singer, and Shieber]{vig2020causalmediationanalysisinterpreting}
Jesse Vig, Sebastian Gehrmann, Yonatan Belinkov, Sharon Qian, Daniel Nevo, Yaron Singer, and Stuart Shieber.
\newblock Investigating gender bias in language models using causal mediation analysis.
\newblock In H.~Larochelle, M.~Ranzato, R.~Hadsell, M.F. Balcan, and H.~Lin, editors, \emph{Advances in Neural Information Processing Systems}, volume~33, pages 12388--12401. Curran Associates, Inc., 2020.
\newblock URL \url{https://proceedings.neurips.cc/paper_files/paper/2020/file/92650b2e92217715fe312e6fa7b90d82-Paper.pdf}.

\bibitem[Wang et~al.(2024)Wang, Hoogland, van Wingerden, Furman, and Murfet]{wang2024differentiationspecializationattentionheads}
George Wang, Jesse Hoogland, Stan van Wingerden, Zach Furman, and Daniel Murfet.
\newblock Differentiation and specialization of attention heads via the refined local learning coefficient, 2024.
\newblock URL \url{https://arxiv.org/abs/2410.02984}.

\bibitem[Wang et~al.(2022)Wang, Variengien, Conmy, Shlegeris, and Steinhardt]{wang2022interpretability}
Kevin Wang, Alexandre Variengien, Arthur Conmy, Buck Shlegeris, and Jacob Steinhardt.
\newblock Interpretability in the wild: a circuit for indirect object identification in gpt-2 small.
\newblock \emph{arXiv preprint arXiv:2211.00593}, 2022.

\bibitem[Watanabe(2009)]{watanabe2009algebraic}
Sumio Watanabe.
\newblock \emph{Algebraic geometry and statistical learning theory}, volume~25.
\newblock Cambridge university press, 2009.

\bibitem[Watson(2022)]{watson2022conceptual}
David~S. Watson.
\newblock Conceptual challenges for interpretable machine learning.
\newblock \emph{Synthese}, 200:\penalty0 65, 2022.
\newblock \doi{10.1007/s11229-022-03485-5}.
\newblock URL \url{https://doi.org/10.1007/s11229-022-03485-5}.

\bibitem[Wright and Sharkey(2024)]{wright2024suppression}
Benjamin Wright and Lee Sharkey.
\newblock Addressing feature suppression in saes, Feb 2024.
\newblock URL \url{https://www.alignmentforum.org/posts/3JuSjTZyMzaSeTxKk/addressing-feature-suppression-in-saes}.

\bibitem[Yun et~al.(2021)Yun, Chen, Olshausen, and LeCun]{yun2023transformervisualizationdictionarylearning}
Zeyu Yun, Yubei Chen, Bruno Olshausen, and Yann LeCun.
\newblock Transformer visualization via dictionary learning: contextualized embedding as a linear superposition of transformer factors.
\newblock In Eneko Agirre, Marianna Apidianaki, and Ivan Vuli{\'c}, editors, \emph{Proceedings of Deep Learning Inside Out (DeeLIO): The 2nd Workshop on Knowledge Extraction and Integration for Deep Learning Architectures}, pages 1--10, Online, June 2021. Association for Computational Linguistics.
\newblock \doi{10.18653/v1/2021.deelio-1.1}.
\newblock URL \url{https://aclanthology.org/2021.deelio-1.1/}.

\bibitem[Zhang et~al.(2021)Zhang, Ahuja, Xu, Wang, and Courville]{zhang2021subnetworkstructurekeyoutofdistribution}
Dinghuai Zhang, Kartik Ahuja, Yilun Xu, Yisen Wang, and Aaron Courville.
\newblock Can subnetwork structure be the key to out-of-distribution generalization?, 2021.
\newblock URL \url{https://arxiv.org/abs/2106.02890}.

\end{thebibliography}
\bibliographystyle{plainnat}
\end{raggedright}

\newpage

\appendix

\section{Appendix}\label{sec:appendix}

\subsection{Architecture of Causal Importance Function MLPs} \label{sec:architecture-causal-importance-mlps}

The MLPs $\gamma^l_c$ that are used in the causal importance function consist of a single hidden layer of GELU neurons with width $d_{\text{gate}}$, followed by a hard sigmoid function:
\begin{equation}
\begin{aligned}
&g^l_c(x)=\sigma_H\left( W^{l,\text{gate out}}_{c}\text{GELU}\left(W^{l, \text{gate in}}_{c} h^l_c(x)+b^{l, \text{gate in}}_{c}\right)+b^{l, \text{gate out}}_{c}\right)\\
&\sigma_H(x):=\begin{cases}
0 &x\leq 0\\
x &0\leq x\leq 1\\
1 &1\leq x
\end{cases}
\end{aligned}
\end{equation}
Here, $h^l_c(x):=\sum_j V^l_{c,j} a^l_j(x)$ is the inner activation of the component and  $W^{l,\text{gate in}}_c\in\mathbb{R}^{d_{\text{gate}}\times 1} ,W^{l, \text{gate out}}\in\mathbb{R}^{1\times d_{\text{gate}}}, b^{l, \text{gate in}}\in\mathbb{R}^{d_{\text{gate}}}, b^{l, \text{gate out}}\in\mathbb{R}$ are trainable parameters. 
Note that there is no sum over $c$ in the above expression: Every subcomponent has its own separate causal importance MLP. This keeps the computational costs of training the gates low compared to the cost of training the subcomponents themselves.

It is important to note that this choice of causal importance function is only one of many possibilities. We chose it for its relative simplicity and low cost. In theory, SPD should be compatible with any method of predicting causal importance values $g^l_c(x)$ for the subcomponents. This is somewhat in contrast to sparse dictionary learning methods, where using arbitrarily expressive nonlinearities and optimization methods to determine dictionary activations raises concerns about whether a `feature' is really represented by the network if it can only be identified in neural activations using a very complex nonlinear function versus a simple thresholded-linear function. See e.g. \citet{bricken2023monosemanticity} for discussion.

\subsection{Avoiding dead gradients with leaky hard sigmoids} \label{sec:avoiding-dead-gradients}
The flat regions in a hard sigmoid function can lead to dead gradients for inputs below $0$ or above $1$. To avoid this, we use leaky hard sigmoids instead, which introduce a small non-zero slope below $0$ or above $1$. Specifically, we use \emph{lower-leaky} hard sigmoids $\sigma_{H,\text{lower}}(x)$ with slope $0.01$ below $0$ for the gates used in the forward pass for the $\mathcal{L}_{\text{stochastic-recon}}$ and $\mathcal{L}_{\text{stochastic-recon-layerwise}}$ losses. And we use \emph{upper-leaky} hard sigmoids $\sigma_{H,\text{upper}}(x)$ with slope $0.01$ above $1$ in the causal importance loss $\mathcal{L}_{\text{importance-minimality}}$ :
\begin{equation}
\begin{aligned}
&\sigma_{H,\text{lower}}(x):=\begin{cases}
0.01x &x\leq 0\\
x &0\leq x\leq 1\\
1 &1\leq x
\end{cases}\\
&\sigma_{H,\text{upper}}(x):=\begin{cases}
0 &x\leq 0\\
x &0\leq x\leq 1\\
1+0.01(x-1) &1\leq x
\end{cases}
\end{aligned}
\end{equation}
We use the lower leaky hard sigmoid for the forward pass because we should usually be able to scale a subcomponent that does not influence the output of the model below zero, but we cannot scale a subcomponent that does influence the output of the model above $1$. So we can use masks smaller than zero in the forward pass, but not masks greater than one.

We use the upper leaky hard sigmoid in the importance loss because the causal importance values cannot be allowed to take negative values, else the training would not be incentivized to sparsify them. But there is no issue with allowing causal importance function outputs to be greater than $1.0$ when computing the importance loss. 

\subsection{Heuristics for Hyperparameter Selection} \label{sec:hyperparam-heuristics}
Here we list some heuristics for how to select hyperparameters to balance the different loss terms of SPD in Equation \ref{eq:full_loss}. In particular, we focus on how to determine the appropriate trade-off between the stochastic reconstruction losses $\mathcal{L}_{\text{stochastic-recon}}$ and $\mathcal{L}_{\text{stochastic-recon-layerwise}}$ (controlled by $\beta_1, \beta_2$) and importance minimality loss $\mathcal{L}_{\text{importance-minimality}}$ (controlled by $\beta_3$).

\begin{itemize}
    \item {Negligible performance loss: The performance difference between the SPD model and the original model on the training dataset should be small enough to be mostly negligible. Quantitatively, one might want to judge how large the performance drop is based on LM scaling curves, as \citet{gao2024scalingevaluatingsparseautoencoders} suggested for judging the quality of SAE reconstructions.}
    \item{Noise from superposition: Inactive mechanisms in superposition can still contribute to the model output through small interference terms \citep{hänni2024mathematicalmodelscomputationsuperposition}. We can estimate the expected size of such terms, and ensure that $\mathcal{L}_{\text{stochastic-recon}}, \mathcal{L}_{\text{stochastic-recon-layerwise}}$ are no larger than what could plausibly be caused by such noise terms.}
    \item{Recovering known mechanisms: If some of the ground truth mechanisms in the target model are already known, we can restrict hyperparameters such that they recover those mechanisms. For example, in a language model, the embedding matrix mechanisms are usually known: Each vocabulary element should be assigned one mechanism. If none of a model’s mechanisms are known to start with, we could insert known mechanisms into it. For example, one might insert an identity matrix at some layer in an LLM and check whether the SPD decomposition recovers it as a single high-rank component, as in the TMS-with-identity model in Section \ref{sec:tms_identity}.}
    \item{Other sanity checks: The decomposition should pass other sanity checks. For example, a model parametrized by the sum of unmasked subcomponents should recover the performance of the target model. And at least some of the causal importance values should take values of $1$ for most inputs; if they do not, then it is likely that the importance minimality loss coefficient $\beta_3$ is too high.}
\end{itemize}

\subsection{Training details and hyperparameters}\label{app:training-details}

\subsubsection{Toy models of superposition (TMS)}\label{app:training-details-tms}

\paragraph{Target model training}
All target models were trained for 10k steps using the AdamW optimizer \citep{loshchilov2019adamw} with weight decay 0.01 and constant learning rate $5\times 10^{-3}$. The dataset uses input features sampled from $[0,1]$ uniformly with probability 0.05, and 0 otherwise.
\begin{itemize}
    \item $\text{TMS}_{5-2}$ and $\text{TMS}_{5-2+\text{ID}}$: batch size 1024
    \item $\text{TMS}_{40-10}$ and $\text{TMS}_{40-10+\text{ID}}$: batch size 8192
\end{itemize}

\paragraph{SPD training: Common hyperparameters}
\begin{itemize}
    \item Optimizer: Adam with max learning rate $1\times 10^{-3}$ and cosine learning rate schedule
    \item Training: 40k steps, batch size 4096
    \item Data distribution: same as target model (feature probability 0.05)
    \item Stochastic sampling: $S=1$ for $\mathcal{L}_{\text{stochastic-recon}}$ and $\mathcal{L}_{\text{stochastic-recon-layerwise}}$
    \item Loss coefficients: $\mathcal{L}_{\text{faithfulness}}=1$, $\mathcal{L}_{\text{stochastic-recon}}=1$, $\mathcal{L}_{\text{stochastic-recon-layerwise}}=1$
    \item Causal importance functions: One MLP per subcomponent, each with one hidden layer of $d_{\text{gate}}=16$ $\text{GELU}$ neurons.
\end{itemize}

\paragraph{SPD training: Model-specific hyperparameters}
\begin{itemize}
    \item $\text{TMS}_{5-2}$ and $\text{TMS}_{5-2+\text{ID}}$: $\mathcal{L}_{\text{importance-minimality}}$ coefficient $3\times10^{-3}$, $p=1$
    \item $\text{TMS}_{40-10}$ and $\text{TMS}_{40-10+\text{ID}}$: $\mathcal{L}_{\text{importance-minimality}}$ coefficient $1\times10^{-4}$, $p=2$ (we expect that one could obtain similar results in these toy models with many different settings for $p$)
\end{itemize}

\subsubsection{Toy models of Compressed computation and Cross-Layer Distributed Representation}\label{app:training-details-resid-mlp}

\paragraph{Model architectures}
\begin{itemize}
    \item 1-layer and 2-layer residual MLPs: 100 input features, embedding dimension 1000, 50 MLP neurons total (25 per layer for 2-layer)
    \item 3-layer residual MLP: 102 input features, embedding dimension 1000, 51 MLP neurons total (17 per layer)
\end{itemize}

\paragraph{Target model training}
All models trained using AdamW with weight decay 0.01, max learning rate $3\times10^{-3}$ with cosine decay, batch size 2048. The dataset uses input features sampled from $[-1,1]$ uniformly with probability 0.01, and 0 otherwise.

\paragraph{SPD training: Common hyperparameters}
\begin{itemize}
    \item Optimizer: Adam with constant learning rate
    \item Batch size: 2048
    \item Data distribution: same as target model (feature probability 0.01)
    \item Stochastic sampling: $S=1$ for both stochastic losses
    \item Loss coefficients: $\mathcal{L}_{\text{stochastic-recon}}=1$, $\mathcal{L}_{\text{stochastic-recon-layerwise}}=1$
    \item $p=2$ for $\mathcal{L}_{\text{importance-minimality}}$
\end{itemize}

\paragraph{SPD training: Model-specific hyperparameters}
\begin{itemize}
    \item 1-layer residual MLP: learning rate $2\times10^{-3}$, $\mathcal{L}_{\text{importance-minimality}}$ coefficient $1\times10^{-5}$, $C=100$ initial subcomponents, 30k training steps, causal importance function with $d_{\text{gate}}=16$ hidden neurons per subcomponent
    \item 2-layer residual MLP: learning rate $1\times10^{-3}$, $\mathcal{L}_{\text{importance-minimality}}$ coefficient $1\times10^{-5}$, $C=400$ initial subcomponents, 50k training steps, causal importance function with $d_{\text{gate}}=16$ hidden $\text{GELU}$ neurons per subcomponent  
    \item 3-layer residual MLP: learning rate $1\times10^{-3}$, $\mathcal{L}_{\text{importance-minimality}}$ coefficient $0.5\times10^{-5}$, $C=500$ initial subcomponents, 200k training steps (converges around 70k), causal importance function with $d_{\text{gate}}=128$ hidden $\text{GELU}$ neurons per subcomponent
\end{itemize}

The hyperparameters can also be found in the WandB report \href{https://wandb.ai/goodfire/spd-tms/reports/SPD-paper-report--VmlldzoxMzE3NzU0MQ?accessToken=427spmsbxig5cyp4jsprg9p183tysclk7ttzyxjlsiwafh8badzlpgxcvopsormm}{here}.

\subsection{Supplementary figures}\label{sec:suppl-figs}
\begin{figure}
    \centering
    \includegraphics[width=1.0\linewidth]{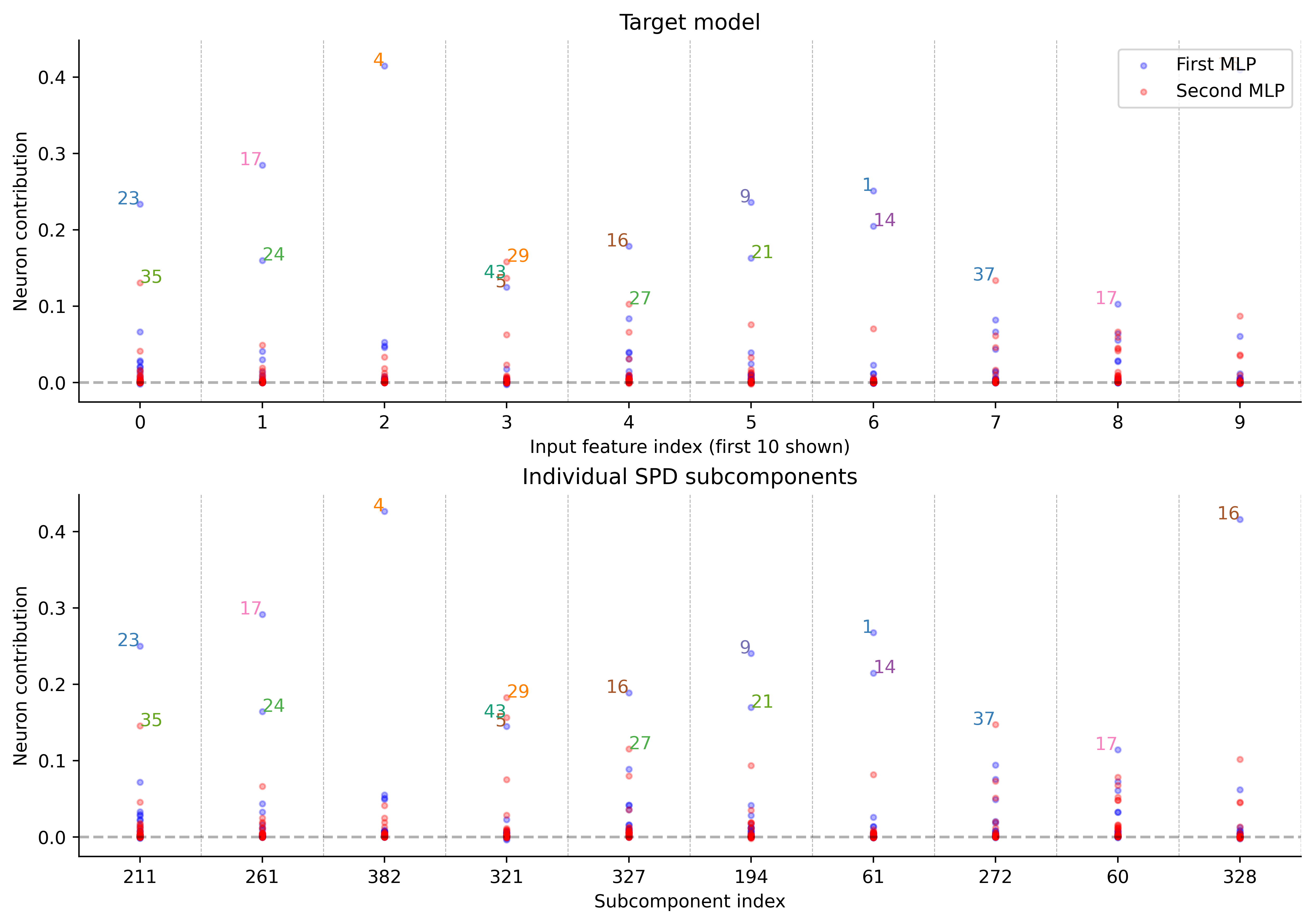}
    \caption{Toy Model of Distributed Representations (Two Layers): Similarity between target model weights and SPD model components for the first $10$ input feature dimensions in a $2$-layer residual MLP. \textbf{Top}: Neuron contributions measured by $(W_E W_\text{IN}) \odot (W_\text{OUT} W_U)$, where $\odot$ is an element-wise product and $W_\text{IN}$ and $W_\text{OUT}$ are the MLP input and output matrices in both layers concatenated together.  \textbf{Bottom}: Neuron contributions for the learned parameter components, measured by $\max_m [({W_U}_{[i,:]} U^\text{OUT} V^\text{OUT}) \odot (U^\text{IN}_{[:,m]} V^\text{IN}_{[m,:]} {W_E}_{[:,i]})]$  for each feature index $i\in[0,9]$. The neurons are numbered based on their raw position in the network, with neurons $0$ to $24$ in the first layer and neurons $25$ to $49$ in the second layer.}
    \label{fig:resid-mlp-weights-2layers}
\end{figure}

\begin{figure}
    \centering
    \includegraphics[width=1.0\linewidth]{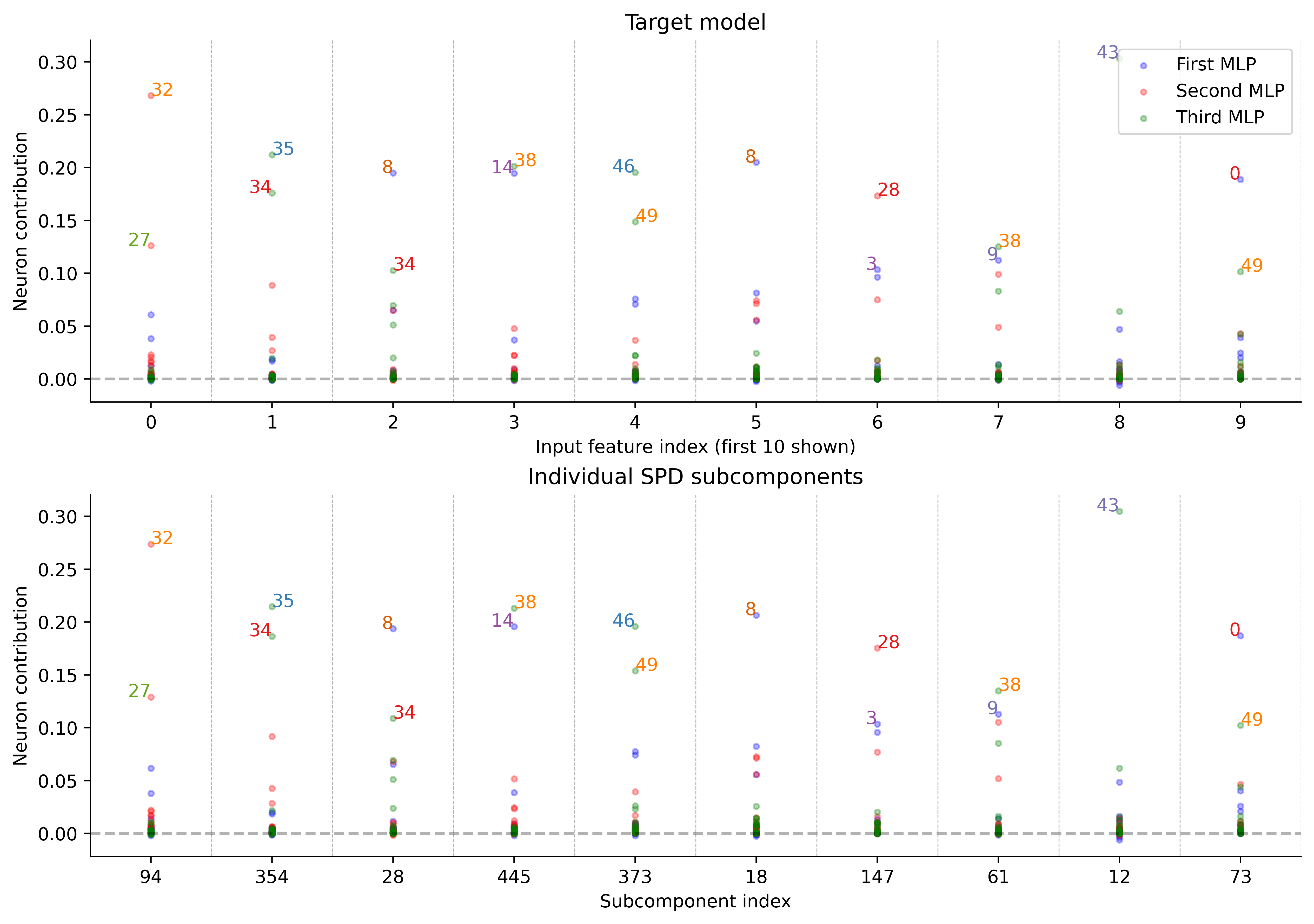}
    \caption{Toy Model of Distributed Representations (Three Layers): Similarity between target model weights and SPD model components for the first $10$ input feature dimensions in a $3$-layer residual MLP. \textbf{Top}: Neuron contributions measured by $(W_E W_\text{IN}) \odot (W_\text{OUT} W_U)$ where $\odot$ is an element-wise product and $W_\text{IN}$ and $W_\text{OUT}$ are the MLP input and output matrices in all three layers concatenated together.  \textbf{Bottom}: Neuron contributions for the learned parameter components, measured by $\max_m [({W_U}_{[i,:]} U^\text{OUT} V^\text{OUT}) \odot (U^\text{IN}_{[:,m]} V^\text{IN}_{[m,:]} {W_E}_{[:,i]})]$ for each feature index $i\in[0,9]$. The neurons are numbered based on their raw position in the network, with neurons $0$ to $16$ in the first layer, $17$ to $33$ in the second layer and neurons $34$ to $50$ in the third layer.}
    \label{fig:resid-mlp-weights-3layers}
\end{figure}

\clearpage
\subsection{SPD Pseudocode}\label{app:pseudocode}
\begin{algorithm}[H]
\caption{Stochastic Parameter Decomposition (SPD)}
\label{alg:spd}
\begin{algorithmic}[1]
\Require Target model $f(\cdot, W)$ with parameters $W=\{W^l\}_{l=1}^L$ to be decomposed.
\Require Dataset $\mathcal{D}$.
\Require $C$ subcomponents per layer.
\Require Loss coefficients $\beta_1, \beta_2, \beta_3$.
\Require Causal Importance Minimality loss p-norm $p > 0$.
\Require Number of mask samples $S \ge 1$.

\Ensure Learned subcomponents $\{U^l, V^l\}_{l=1}^L$ and parameters for causal importance MLP $\{\Gamma^l\}_{l=1}^L$.

\State Initialize subcomponents $U^l \in \mathbb{R}^{d_{out}\times C}$, $V^l \in \mathbb{R}^{C\times d_{in}}$ for each layer $l$.
\State Initialize parameters for causal importance MLPs $\Gamma^l$.
\State Initialize an optimizer for all trainable parameters ($\{U^l, V^l, \Gamma^l\}_{l=1}^L$).

\For{each training step}
    \State Sample a data batch $X = \{x_1, \dots, x_B\}$ from $\mathcal{D}$.
    \State Compute target model outputs $Y_{\text{target}}$ and pre-weight activations $\{a^l\}_{l=1}^L$ for batch $X$.

    \State $\mathcal{L}_{\text{faithfulness}} \gets \frac{1}{N} \sum_{l=1}^L \|W^l - U^l V^l\|_F^2$

    \For{each layer $l=1,\dots,L$}
        \State $h^l \gets  V^l a^l$ \textit{\quad // Inner activations}
        \State $G^l_{\text{raw}} \gets \Gamma^l(h^l)$ \textit{\quad // Raw causal importance MLP outputs}
    \EndFor
    \State $\mathcal{L}_{\text{importance-minimality}} \gets \frac{1}{B} \sum_{b=1}^B \sum_{l=1}^L \sum_{c=1}^C |\sigma_{H,\text{upper}}(G^{l}_{\text{raw}, b, c})|^p$

    \State $\mathcal{L}_{\text{stochastic-recon}} \gets 0$
    \State $\mathcal{L}_{\text{stochastic-recon-layerwise}} \gets 0$
    \State Let $G^l = \sigma_{H,\text{lower}}(G^l_{\text{raw}})$ for all $l$.

    \For{$s=1, \dots, S$}
        \State Sample $R_s^l \sim \mathcal{U}(0,1)^{B \times C}$ for each layer $l$.
        \State Compute masks $M_s^l \gets G^l + (1 - G^l) \odot R_s^l$.

        \State Construct masked weights $\{W'^{(s)}_b\}_{b=1}^B$ with $W'^{(s), l}_b = U^l \cdot \text{Diag}(M_{s, b}^{l}) \cdot V^l$.
        \State $Y_{\text{masked}} \gets f(X, W'^{(s)})$
        \State $\mathcal{L}_{\text{stochastic-recon}} \gets \mathcal{L}_{\text{stochastic-recon}} + D(Y_{\text{masked}}, Y_{\text{target}})$

        \For{layer $l'=1, \dots, L$}
            \State Construct layerwise masked weights $\{W'^{(s,l')}_b\}_{b=1}^B$ as follows:
            \Statex \hspace{6em} $W'^{(s,l'), l}_b = U^l \cdot \text{Diag}(M_{s,b}^{l} \text{ if } l = l' \text{ else } \mathbf{1}) \cdot V^l$
            \State $Y_{\text{masked-layerwise}} \gets f(X, W'^{(s, l')})$
            \State $\mathcal{L}_{\text{stochastic-recon-layerwise}} \gets \mathcal{L}_{\text{stochastic-recon-layerwise}} + D(Y_{\text{masked-layerwise}}, Y_{\text{target}})$
        \EndFor
    \EndFor
    \State Normalize $\mathcal{L}_{\text{stochastic-recon}} \gets \mathcal{L}_{\text{stochastic-recon}} / S$
    \State Normalize $\mathcal{L}_{\text{stochastic-recon-layerwise}} \gets \mathcal{L}_{\text{stochastic-recon-layerwise}} / (S \cdot L)$

    \State $\mathcal{L}_{\text{SPD}} \gets \mathcal{L}_{\text{faithfulness}} + \beta_1 \mathcal{L}_{\text{stochastic-recon}} + \beta_2 \mathcal{L}_{\text{stochastic-recon-layerwise}} + \beta_3 \mathcal{L}_{\text{importance-minimality}}$
    \State Update parameters of $\{U^l, V^l, \Gamma^l\}_{l=1}^L$ using gradients of $\mathcal{L}_{\text{SPD}}$.
\EndFor
\end{algorithmic}
\end{algorithm}

\end{document}